%% file: main.tex
\documentclass[10pt,twocolumn,letterpaper]{article}

\usepackage{cvpr}              %
\usepackage{svg}
\usepackage{booktabs} %
\usepackage{xcolor}   %
\usepackage{pifont}   %
\usepackage{amssymb}  %
\usepackage{multirow}
\usepackage{rotating}

\input{preamble}

\definecolor{cvprblue}{rgb}{0.21,0.49,0.74}
\usepackage[pagebackref,breaklinks,colorlinks,allcolors=cvprblue]{hyperref}
\usepackage{bbm}
\usepackage{lipsum}
\usepackage{ifthen}

\def\paperID{16} %
\def\confName{CVPR}
\def\confYear{2026}

\title{DINO-QPM: Adapting Visual Foundation Models for Globally Interpretable Image Classification}

\author{Robert Zimmermann$^*$ \quad Thomas Norrenbrock$^*$ \quad Bodo Rosenhahn\\
Institute for Information Processing, L3S - Leibniz University Hannover\\
{\tt\small \{zimmerro, norrenbr, rosenhahn\}@tnt.uni-hannover.de}
}

\begin{document}
\twocolumn[{%
\renewcommand\twocolumn[1][]{#1}%
\maketitle

\begin{center}
    \centering
    \includegraphics[width=0.71\textwidth]{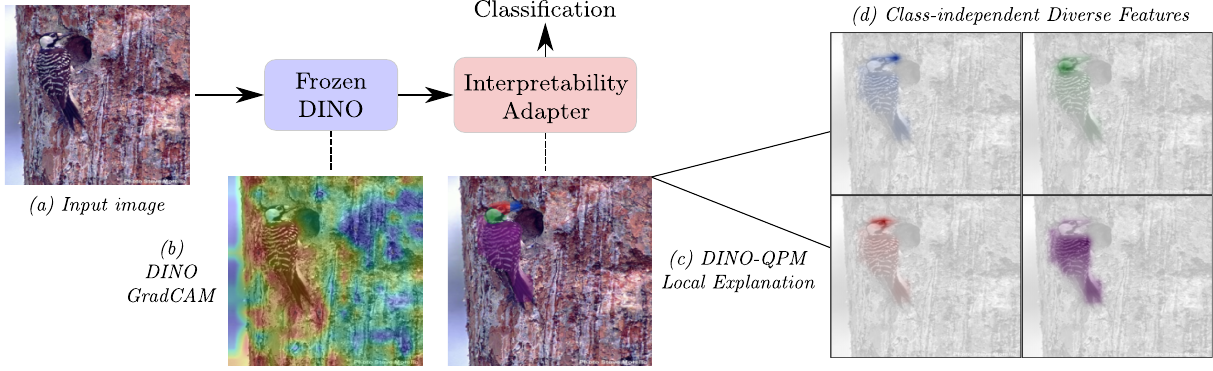}
    \captionof{figure}{Overview of our proposed DINO-QPM. The pipeline processes the (a) input image using the frozen backbone to produce patch embeddings, which are transformed by the interpretability adapter to obtain a globally interpretable image classification. We compare the diffuse saliency map of (b) DINO GradCAM, extracted from a linear probed DINO model, with our (c) DINO-QPM local explanation. The local explanation can be further decomposed into its (d) class-independent diverse features. Compared to the baseline, we observe a drastic increase in localisation quality, showcasing how our interpretability adapter successfully isolates semantically meaningful features.}
    \label{fig:main_intro_fig}
\end{center}%
}]

\input{sec/0_abstract}    
\input{sec/1_intro}
\input{sec/2_related_work}
\input{sec/3_fundamentals}

\input{sec/4_method}
\input{sec/5_experiments}

\input{sec/6_conclusion}
\newpage

\input{sec/7_acknowledgements}

\small
\bibliographystyle{ieeenat_fullname}
\bibliography{main}

\input{sec/X_suppl}

\end{document}

%% file: preamble.tex
\newcommand\blfootnote[1]{%
  \begingroup
  \renewcommand\thefootnote{}\footnote{#1}%
  \addtocounter{footnote}{-1}%
  \endgroup
}

%% file: sec/0_abstract.tex
\begin{abstract}
\blfootnote{$^*$ Indicates equal contribution.}
Although visual foundation models like DINOv2 provide state-of-the-art performance as feature extractors, their complex, high-dimensional representations create substantial hurdles for interpretability. This work proposes DINO-QPM, which converts these powerful but entangled features into contrastive, class-independent representations that are interpretable by humans. DINO-QPM is a lightweight interpretability adapter that pursues globally interpretable image classification, adapting the Quadratic Programming Enhanced Model (QPM) to operate on strictly frozen DINO backbones. While classification with visual foundation models typically relies on the \texttt{CLS} token, we deliberately diverge from this standard. By leveraging average-pooling, we directly connect the patch embeddings to the model's features and therefore enable spatial localisation of DINO-QPM's globally interpretable features within the input space. Furthermore, we apply a sparsity loss to minimise spatial scatter and background noise, ensuring that explanations are grounded in relevant object parts. With DINO-QPM we make the level of interpretability of QPM available as an adapter while exceeding the accuracy of DINOv2 linear probe. Evaluated through an introduced Plausibility metric and other interpretability metrics, extensive experiments demonstrate that DINO-QPM is superior to other applicable methods for frozen visual foundation models in both classification accuracy and explanation quality. 
\end{abstract}

%% file: sec/1_intro.tex
\section{Introduction}
\label{sec:intro}
Visual foundation models such as DINOv2 \cite{oquabDINOv2LearningRobust2024} have shown great performance as powerful, general-purpose feature extractors across various image analysis benchmarks \cite{oquabDINOv2LearningRobust2024, Zhang2023, Izquierdo2023, Cui2024, kaiseruncertainsam}. However, deploying these models in safety-critical domains requires a high degree of interpretability, which remains a significant challenge due to their complex, opaque architectures.

Thus interpretable-by-design approaches are getting more popular for such applications \cite{rudin2019stop}. Inspired by human cognitive processes \cite{rosch1978principles}, one line of work computes the similarity to so-called prototypes to obtain an interpretable classification \cite{chenThisLooksThat2019, nautaNeuralPrototypeTrees2021, rymarczykProtoPSharePrototypeSharing2021}. \citet{turbeProtoSViTVisualFoundation2024}, \citet{zhuInterpretableImageClassification2025}, \citet{maInterpretableImageClassification2024} and \citet{turbé2025tellwhyvisualfoundation} apply this idea to visual foundation models. However, prototypical model have a deceiving interpretability as their similarity is not restricted to be similar to humans \cite{kim2022hiveevaluatinghumaninterpretability, looks_like_that_does_it, Baniecki_2025}. Therefore several other approaches utilise sparse, low-dimensional, and quantised decision layers to enforce compact class representations \cite{norrenbrockTake5Interpretable2023, norrenbrockQSENNQuantizedSelfExplaining2024, norrenbrockQPMDiscreteOptimization2025, norrenbrockCHiQPMCalibratedHierarchical2025}. Unlike local interpretability methods that only explain individual predictions, these approaches aim for global interpretability, providing a holistic, transparent view of the model's entire decision-making process and how it defines classes across the dataset. This leads to diverse \cite{jakkola_diverse}, contrastive \cite{Lipton_1990}, general and compact \cite{readExplanatoryCoherenceSocial1993} feature representations ideally suited for generating human-interpretable explanations \cite{MILLER20191}. 

Many of the aforementioned methods are fully trained end-to-end and therefore require massive resources for training purposes. While recent works have begun exploring interpretability techniques like post-hoc concept mapping \cite{yuksekgonulPosthocConceptBottleneck2023, oikarinenLabelFreeConceptBottleneck2023, yangLanguageBottleLanguage2023} on top of frozen backbones, these approaches struggle to reach a competitive level of accuracy. To address these limitations, our approach aligns with the objective of models such as QPM \cite{norrenbrockQPMDiscreteOptimization2025} and ChiQPM \cite{norrenbrockCHiQPMCalibratedHierarchical2025}, aiming to represent classes through general, diverse and contrastive features. Translating the  mathematically constrained compactness of sparse, quantised decision layers, like those used in QPM \cite{norrenbrockQPMDiscreteOptimization2025}, to frozen visual foundation models for inherently interpretable image classification remains an open problem. 

In this work, we address this problem by applying the Quadratic Programming Enhanced Model (QPM) \cite{norrenbrockQPMDiscreteOptimization2025} to the high-dimensional representations of DINOv2 and building a lightweight interpretability adapter on top of its frozen features, as suggested in \citet{simeoniDINOv32025}. The adapter transforms DINOv2's powerful, entangled representations using a sparse feature assignment into diverse, class-independent and contrastive features, yielding a globally interpretable solution for image classification. 

After applying an MLP to the patch embeddings of the frozen backbone, we use average-pooling to obtain a feature vector, which is inherently connected to the problem-specific feature maps returned by the MLP. Although the standard choice in classification literature is to, at least partially, use the \texttt{CLS} token \cite{oquabDINOv2LearningRobust2024, chenContextAutoencoderSelfSupervised2023}, this direct connection to the feature maps enables high-fidelity spatial localisation of features in the image. 

\begin{figure}[t]
	\centering
	\includegraphics[width=0.49\textwidth]{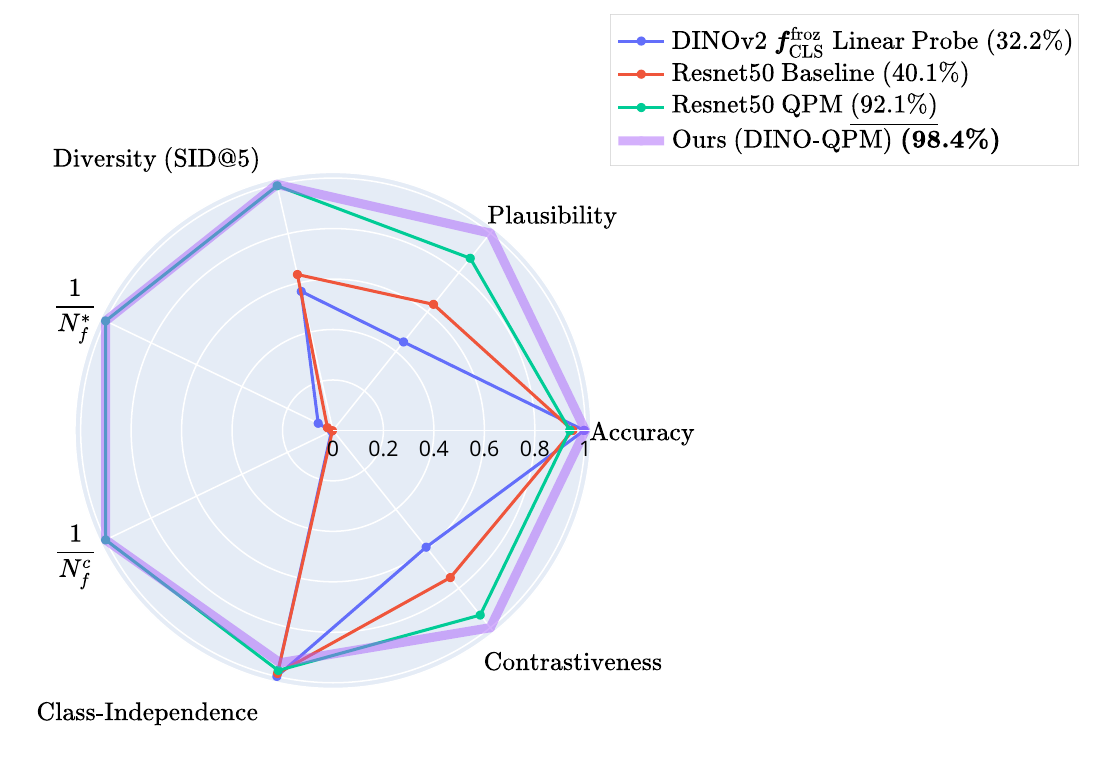}
	\caption{Radar Plot demonstrating the quality of DINO-QPM in a set of interpretability metrics and accuracy compared to non-frozen QPM and the corresponding baselines. DINO-QPM outperforms each of them reaching $98.4 \%$ of the maximal score calculated as the fraction of area of the heptagon reached by the respective model. We present rigorous insights in \cref{sec:experiments} including a more detailed presentation of the interpretability metrics. }
	\label{fig:radar_plot}
\end{figure}

Inspired by the pointing game \cite{zhang2016top}, we introduce a Plausibility metric, which measures the fraction of the cumulative feature map activation that falls within the object boundaries. We are able to quantify that DINO-QPM's interpretable features localise consistently on the relevant object parts, while the saliency map of a linear probe on the frozen features lacks in Plausibility, as  visualised in \cref{fig:main_intro_fig} and quantitatively shown in \cref{fig:radar_plot}. We further apply a sparsity loss to enhance spatial precision of our model. The sparsity loss effectively minimises spatial scatter and background noise, which ensures that model explanations are strictly grounded in relevant object parts. 

Extensive validation across multiple datasets and backbones confirms that DINO-QPM outperforms state-of-the-art interpretable methods applied to visual foundation models in terms of both classification accuracy and explanation quality. To facilitate reproducibility and future research, the code is available at \url{https://github.com/RobertZimm/DINO-QPM}.

The main contributions of this work are:

\begin{itemize}
    \item \textbf{Lightweight Interpretability Adapter for Frozen Backbones:} Our proposed lightweight interpretability adapter (DINO-QPM) is designed to function with frozen self-supervised backbones, such as DINOv2. This design facilitates inherently interpretable image classification without the need for full model fine-tuning or high computational overhead, delivering state-of-the-art interpretability on top of frozen visual foundation models while maintaining exceptional accuracy.
    
    \item \textbf{Spatial Localisation through Token Representations:} DINO-QPM leverages average-pooling across tokens to enable the spatial localisation of its features in the input space. This allows the generation of high-fidelity saliency maps, while beating the linear probe in accuracy and outperforming the dense average-pooled variant by more than $10 \%$ on CUB-2011 \cite{wahCaltechUCSDBirds2002011Dataset2011}.
    
    \item \textbf{Enhanced Plausibility via Sparsity Loss:} In contrast to the initial feature maps from a DINOv2 linear probe, DINO-QPM has an exceptional localisation ability, which is further enhanced using a sparsity loss. We quantify this via our introduced Plausibility metric.
\end{itemize}

%% file: sec/2_related_work.tex
\section{Related Work}
\label{sec:related_work}
Initial approaches to interpreting visual foundation models rely heavily on visualising attention maps and feature projections. For instance, \citet{dosovitskiyImageWorth16x162021} and \citet{caronDinoV1EmergingProperties2021} visualise the last self-attention layer of the CLS token to understand feature extraction. Similarly, for DINOv2, \citet{oquabDINOv2LearningRobust2024} visualise patch embeddings by filtering the foreground using the first PCA principal component, then transforming the embeddings via a second PCA to display their primary components as RGB channels. While these visualisations suggest that foundation models inherently learn well-localising features without explicit supervision \cite{caronDinoV1EmergingProperties2021}, they serve primarily as qualitative observations rather than rigorous explanations.

The assumption that attention maps suffice for credible explanations is heavily criticised in NLP \cite{jainAttentionNotExplanation2019} and vision \cite{cheferTransformerInterpretabilityAttention2021, brockiClassDiscriminativeAttentionMaps2024}. Attention maps are inherently independent of downstream tasks, meaning they often neglect crucial information required for classification \cite{cheferTransformerInterpretabilityAttention2021, brockiClassDiscriminativeAttentionMaps2024}. Derived solely from query-key products, they ignore the highly influential value vectors and MLP blocks \cite{cheferTransformerInterpretabilityAttention2021, chungEvaluatingVisualExplanations2025}.

While more advanced post-hoc methods have emerged to aggregate signals across the entire architecture—including Gradient Attention Rollout \cite{abnarQuantifyingAttentionFlow2020, kashefiExplainabilityVisionTransformers2023}, LRP \cite{cheferTransformerInterpretabilityAttention2021}, CDAM \cite{brockiClassDiscriminativeAttentionMaps2024}, and ViT-Shapley \cite{covertLearningEstimateShapley2023}—these techniques remain external approximations rather than inherent, faithful reflections of the model's decision-making process. 

To overcome the limitations of post-hoc explanations, interpretable-by-design (ad-hoc) approaches integrate the explanation directly into the model's decision process. Current approaches utilise sparse \cite{glandorf2023hypersparse, glandorf2025pruning, Bod2023a}, low-dimensional, and quantised decision layers to inherently increase interpretability \cite{norrenbrockTake5Interpretable2023, norrenbrockQSENNQuantizedSelfExplaining2024, norrenbrockQPMDiscreteOptimization2025, norrenbrockCHiQPMCalibratedHierarchical2025}. Translating these inherent interpretability mechanisms to visual foundation models, however, presents unique challenges. Existing ad-hoc architectures for vision transformers include IA-RED$^{2}$ \cite{panIARED^2InterpretabilityAwareRedundancy2021}, the B-Cos alignment approach \cite{bohleBcosNetworksAlignment2022, bohleBcosAlignmentInherently2024}, and prototype-based methods like ProtoViT \cite{maInterpretableImageClassification2024}. While recent works like \citet{zhuInterpretableImageClassification2025} successfully achieve part-based interpretability, they depend on fine-tuning the backbone to enforce prototype clustering. 

Post-hoc Concept Bottleneck Models (CBMs) maintain a completely frozen backbone, but they rely on textual concept supervision rather than providing direct spatial localisation \cite{yuksekgonulPosthocConceptBottleneck2023}. While recent advancements in Post-hoc CBMs have automated concept selection via Large Language Models \cite{oikarinenLabelFreeConceptBottleneck2023, sunConceptBottleneckLarge2025, zhaoPartiallySharedConcept2025} or secondary segmentation networks \cite{prasseDCBMDataEfficientVisual2025}, they still largely depend on external textual supervision or auxiliary trained modules. \\
To the best of our knowledge, DINO-QPM is the first approach to extract sparse, spatially localised, and class-independent part explanations directly from frozen DINOv2 features without requiring external concept banks, language models, or expensive backbone fine-tuning.

%% file: sec/3_fundamentals.tex
\section{Fundamentals}
\subsection{QPM}
\label{subsec:qpm}
\citet{norrenbrockQPMDiscreteOptimization2025} introduce the Quadratic Programming Enhanced Model (QPM) as a model which learns globally interpretable class representations.
The QPM architecture is characterised by a decision layer that is both sparse (weight matrix $\boldsymbol{W}$ contains only a few non-zero entries) and low-dimensional.

\begin{figure}[!ht]
    \centering
    \includegraphics[scale=.7]{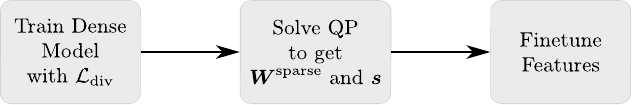}
    \caption{Three-Stage Training Procedure for QPM}
    \label{fig:qpm-pipeline}
\end{figure}

To transform a dense model (non-sparse weights, conventional neural network) into a QPM, the pipeline illustrated in \cref{fig:qpm-pipeline} is employed. To reduce conceptual ambiguity between features, \citet{norrenbrockTake5Interpretable2023} introduced the Feature Diversity Loss, hereafter referred to as $\mathcal{L}_{\text{div}}$ and defined in \cref{app:fdl}. The objective of $\mathcal{L}_{\text{div}}$ is to encourage the representation of distinct, mutually independent concepts within the features, thereby enhancing the degree of model interpretability.

Initially, the dense model is trained using the $\mathcal{L}_{\text{div}}$ loss to ensure the emergence of diverse features. To enforce model sparsity and achieve the required low-dimensional feature space, the framework solves a quadratic program (QP). This QP is used to select $N_f^*$ features from the $N_f$ features of the dense model (where $N_f^* \ll N_f$). The selection is defined by a vector $\boldsymbol{s} \in \{0,1\}^{N_f}$, such that: 

\begin{equation}
    \sum_{d \in \mathcal{F}} s_d = N_f^*
\end{equation}

where $d \in \mathcal{F} = \{1, \ldots, N_f\}$ indexes the set of all features $\mathcal{F}$.
Simultaneously each class $c \in \mathcal{C}$ is assigned a subset of these selected features $d \in \mathcal{F}^*$ consisting of $N_f^{c}$ elements. This mapping is $\boldsymbol{W}^{\text{sparse}} \in \{0,1\}^{N_c \times N_f^*}$, where $W^{\text{sparse}}_{cd}$ is $1$ if class $c$ is assigned feature $d$, and $0$ otherwise. Consequently:

\begin{equation}
    \sum_{d \in \mathcal{F}} W^{\text{sparse}}_{cd} = N_f^{c} \qquad \forall \; c \in \mathcal{C} 
\end{equation}

The assignment process relies on the optimisation of three components defined in \cite{norrenbrockQPMDiscreteOptimization2025}: Maximising the correlation between class $c$ and its assigned feature activations, minimising similarity between selected features $\boldsymbol{s}$, and maximising the bias $b_d$ to prioritise local features. For a comprehensive derivation of the QP objective function, the reader is referred to the original work \cite{norrenbrockQPMDiscreteOptimization2025}. 
Finally the model is retrained with the constraint $\boldsymbol{W} = \boldsymbol{W}^{\text{sparse}}$, considering only the subset of selected features $\mathcal{F}^* \subset \mathcal{F}$ to arrive at the final QPM. When fine-tuning the features adapt to this forced assignment and finally become more interpretable (see \cref{subsec:main_results}). 

\subsection{DINO}
DINO (Self-\textbf{DI}stillation with \textbf{NO} labels), introduced by \citet{caronDinoV1EmergingProperties2021}, is a framework for self-supervised representation learning \cite{hadsellDimensionalityReductionLearning2006, grillBootstrapYourOwn2020, heMomentumContrastUnsupervised2020} of visual embeddings. Unlike traditional contrastive learning approaches \cite{hadsellDimensionalityReductionLearning2006, chenSimpleFrameworkContrastive2020, schroffFaceNetUnifiedEmbedding2015, oordRepresentationLearningContrastive2019}, DINO does not rely on negative samples to distinguish the input from other instances and yet achieves remarkable results in various image analysis benchmarks  \cite{oquabDINOv2LearningRobust2024, Zhang2023, Izquierdo2023, Cui2024}.
Beyond architectural changes, its successor DINOv2 utilises a novel pipeline for large-scale training data curation \cite{oquabDINOv2LearningRobust2024}.
\citet{darcetVisionTransformersNeed2024} investigate a specific challenge exacerbated by the transition to DINOv2, though present in other ViT architectures: outlier tokens (or artifacts) in attention maps characterised by unusually high norms.  \citet{darcetVisionTransformersNeed2024} conclude that the ViT utilises these tokens to store global context in areas of low information density. To mitigate this, the authors propose adding "register tokens", non-spatial tokens similar to the CLS token, to act as a storage for global information. These registers are discarded during downstream analysis tasks \cite{darcetVisionTransformersNeed2024}.

In our experiments (\cref{sec:experiments}), we compare various sizes of DINOv2, both with and without register tokens, demonstrating their significant benefit for DINO-QPM. Further, evaluation is conducted on the original DINO \cite{caronDinoV1EmergingProperties2021}.

%% file: sec/4_method.tex
\section{Method}
\label{sec:method}
\begin{figure}[t]
	\centering
	\includegraphics[width=0.3\textwidth]{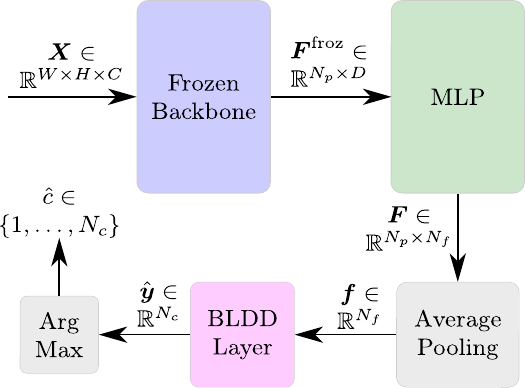}
	\caption{Architecture of our proposed DINO-QPM. Patch embeddings extracted using the frozen backbone are first projected via an MLP into the problem-specific feature space. Subsequently, our BLDD layer performs sparse feature assignment to yield a globally interpretable image classification.}
	\label{fig:Modell-scheme_avg_pooling}
\end{figure}

\begin{figure*}[t]
    \centering
    \includegraphics[width=0.9\textwidth]{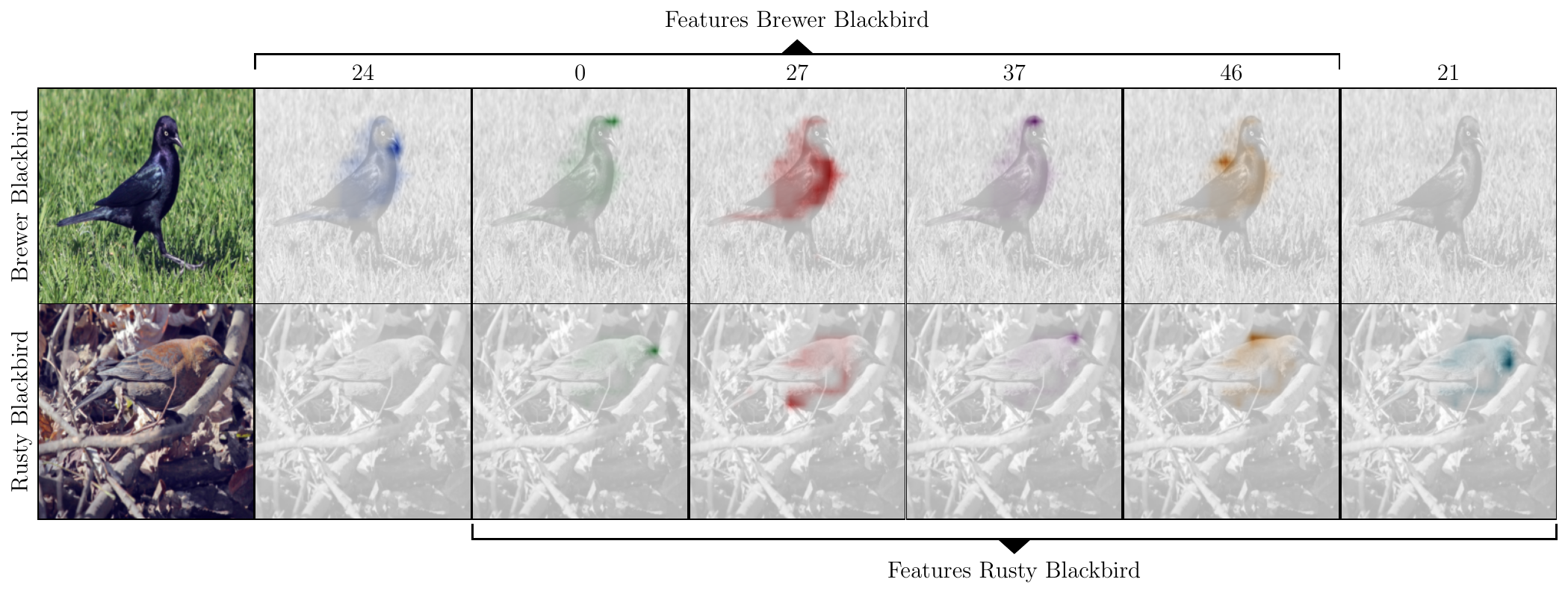}
    \caption{Comparison of a Brewer's Blackbird image with a Rusty Blackbird image. From the selected features $\mathcal{F}^{*}$, $N_f^{\hat{c}}=5$ utilised features were selected for both classes using the QP; the corresponding feature maps from $\boldsymbol{F}$ are visualised as saliency maps. Both classes share 4 out of the 5 features and can thus be distinguished by the non-shared features. Notably, the model differentiates the Brewer's Blackbird using feature 24, which localises the beak. This aligns perfectly with established ornithological expertise, where beak morphology is considered a primary diagnostic trait \cite{savignacCOSEWICAssessmentStatus2006, RustyBlackbirdIdentification}.} 
    \label{fig:example_vis2}
\end{figure*}

Our contribution is DINO-QPM, a model that achieves globally interpretable image classification by enforcing a structured decision process. Given an input image $\boldsymbol{X}$ from the input space $\mathcal{V} = \mathbb{R}^{W \times H \times C}$ where $H$ and $W$ are height and width of the image respectively and $C$ is the number of image channels, a visual foundation model $\Phi$, such as DINOv2, produces a global image representation $\boldsymbol{f}_{\text{CLS}}^{\text{froz}} \in \mathbb{R}^{D}$ and local image representations $\boldsymbol{F}^{\text{froz}} \in \mathbb{R}^{N_p \times D}$. $\boldsymbol{F}^{\text{froz}}$ can be interpreted either as a collection of $D$ spatial feature maps of size $N_p$ or as $N_p$ individual patch representations embedded in $\mathbb{R}^{D}$. The objective is to construct a classifier $g: \mathcal{V} \rightarrow \mathcal{C}$ based on $\boldsymbol{F}^{\text{froz}}$ and $\boldsymbol{f}_{\text{CLS}}^{\text{froz}}$, which maps each input data point $\boldsymbol{X} \in \mathcal{V}$ to a corresponding class $c \in \mathcal{C}$. We introduce the superscript "froz" to indicate the frozen nature of the backbone's output. 

To achieve this, our approach builds upon the QPM framework proposed by \citet{norrenbrockQPMDiscreteOptimization2025} to identify a subset of selected features $\mathcal{F}^{*} \subset \mathcal{F} = \{ 1, \dots, N_f \}$ and assign $N_f^c$ features from this subset to each class. The exact procedure is shown in \cref{fig:Modell-scheme_avg_pooling}. Interestingly, it is sufficient to consider exclusively the frozen local feature maps $\boldsymbol{F}^{\text{froz}} \in \mathbb{R}^{N_p \times D}$ while discarding the global feature vector $\boldsymbol{f}_{\text{CLS}}^{\text{froz}}$, a choice we justify empirically in \cref{subsec:ablations}. This architectural choice is driven by the hypothesis that a global representation transparently built from local evidence is more inherently interpretable than the complex, pre-aggregated representation of the CLS token. By intentionally discarding $\boldsymbol{f}_{\text{CLS}}^{\text{froz}}$, we prevent its internal, opaque aggregation process from introducing features that cannot be inherently localised. 

The purpose of the MLP is to facilitate a task-specific transformation $\text{MLP}: \mathbb{R}^{D} \rightarrow \mathbb{R}^{N_f}$ for the patch representations $\boldsymbol{F} \in \mathbb{R}^{N_p \times N_f}$. This transformation maps the initial $D$-dimensional space to $N_f$ features, which is essential because a sparse Binary Low-Dimensional Decision (BLDD) layer alone provides no capacity for such a transformation. 

The final feature vector $\boldsymbol{f} = \text{AvgPool}\big(\boldsymbol{F}\big) \in \mathbb{R}^{N_f}$ is then constructed as a direct average across the spatial dimensions. Consequently, $\boldsymbol{F}$ contains a saliency map of $N_p$ elements for each feature $d \in \mathcal{F}$. These saliency maps can be upsampled to the original input image resolution for visualisation purposes (see \cref{fig:example_vis2}) and directly highlight where evidence for each feature is found.

Subsequently, the BLDD layer $\text{BLDD}: \mathbb{R}^{N_f} \rightarrow \mathbb{R}^{N_c}$ is applied to the feature vector $\boldsymbol{f}$:

\begin{equation*}
    \text{BLDD}(\boldsymbol{f}) =
    \begin{cases}
        \boldsymbol{W}^{\text{sparse}}\boldsymbol{f}_{(\boldsymbol{s})}, & \text{if fine-tuning} \\
        \boldsymbol{W}\boldsymbol{f}, & \text{otherwise}
    \end{cases}
\end{equation*}

During the dense training phase, the BLDD layer maps the feature vector $\boldsymbol{f}$ to the classification vector $\hat{\boldsymbol{y}}$ using a dense projection matrix $\boldsymbol{W} \in \mathbb{R}^{N_c \times N_f}$. 

In the fine-tuning stage, the selection vector $\boldsymbol{s} \in \{0,1\}^{N_f}$ is used to extract $N_f^{*}$ features from the $N_f$ elements of the feature vector, yielding $\boldsymbol{f}_{(\boldsymbol{s})} \in \mathbb{R}^{N_f^*}$. This subset $\boldsymbol{f}_{(\boldsymbol{s})}$ is then mapped to the classification vector $\hat{\boldsymbol{y}}$ via the sparse projection matrix $\boldsymbol{W}^{\text{sparse}} \in \{ 0, 1 \}^{N_c \times N_f^{*}}$; notably, both $\boldsymbol{s}$ and $\boldsymbol{W}^{\text{sparse}}$ are derived using the QPM (\cref{subsec:qpm}). Specifically, each class $c \in \mathcal{C} = \{ 1, \dots, N_c \}$ is assigned exactly $N_f^{c}$ features. In both cases, the predicted class $\hat{c}$ is determined as the index of the maximum value in $\hat{\boldsymbol{y}}$.

For training purposes we use the exact pipeline described in \cref{subsec:qpm}, also to determine $\boldsymbol{s}$ and $\boldsymbol{W}^{\text{sparse}}$. Besides the Cross-Entropy loss $\mathcal{L}_{\text{CE}}$ and the already introduced $\mathcal{L}_{\text{div}}$, our loss function consists of two L1 sparsity losses, one for the feature vector $\mathcal{L}_{\text{L1-FV}}=\text{Mean}(\text{Abs}(\boldsymbol{f}))$ and $\mathcal{L}_{\text{L1-FM}}=\text{Mean}(\text{Abs}(\boldsymbol{F}))$ for the feature maps which are used to reduce spatial clutter and background noise in the feature maps to significantly increase accuracy and Plausibility. We explicitly show these benefits in \cref{subsec:l1_fm} and \cref{app:losses} respectively. 

%% file: sec/5_experiments.tex
\section{Experiments}
\label{sec:experiments}
We evaluate our proposed DINO-QPM on the problem of fine-grained image classification, in line with previous work on inherently interpretable models \cite{norrenbrockCHiQPMCalibratedHierarchical2025, norrenbrockQPMDiscreteOptimization2025, norrenbrockQSENNQuantizedSelfExplaining2024, norrenbrockTake5Interpretable2023}. 
Stanford Cars \cite{krause3DObjectRepresentations2013} and CUB-2011 \cite{wahCaltechUCSDBirds2002011Dataset2011} are used as datasets, as they are the default for this problem. For CUB-2011 we do not use cropping to the object of interest to demonstrate how DINO-QPM exploits the strong general features of its backbone. Additionally, CUB-2011 offers human annotated masks of the region of interests which enables a quantification of a models Plausibility.
DINOv2 ViT-B/14 \cite{oquabDINOv2LearningRobust2024} with register tokens \cite{darcetVisionTransformersNeed2024} serves as the primary backbone for the various experiments while we also test out different sizes with and without registers as well as DINOv1 \cite{caronDinoV1EmergingProperties2021}. Following the QPM \cite{norrenbrockQPMDiscreteOptimization2025}, $N_f^{*}=50$ features are selected from the initial $N_f$ features, with $N_f^{c}=5$ features assigned to each class, unless otherwise noted. Further details on our implementation can be found in \cref{app:implementation_details}. For all experiments, we train on five different randomly chosen seeds and calculate statistics such as the mean and standard deviation (see \cref{app:detailed_results}).

\subsection{Metrics}
\label{subsec:metrics}
\begin{figure}[t]
    \centering
    \includegraphics[width=0.65\columnwidth]{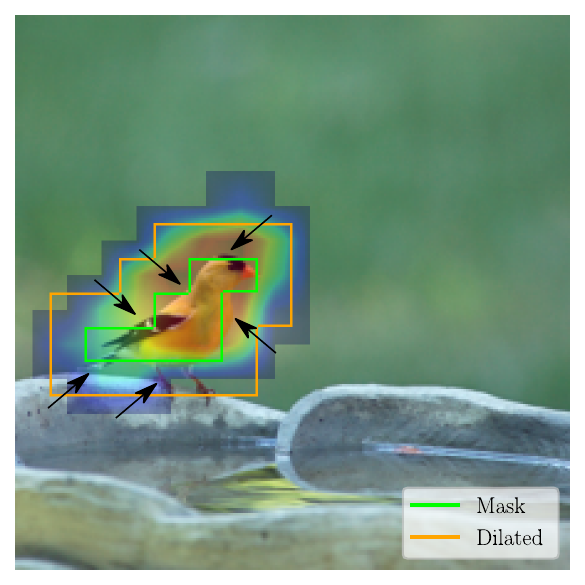}
    \caption{Visualisation of the Plausibility metric on a sample from the CUB2011 dataset (American Goldfinch). The metric quantifies the fraction of the cumulative weighted feature map activation $\widetilde{\boldsymbol{F}}$ that falls within the dilated object segmentation mask $\boldsymbol{M}^{\text{dil}}$. The arrows highlight the relevant features at the objects' edge that are missed by the non-dilated mask.}
    \label{fig:o_mean}
\end{figure}

We evaluate our model following QPM \cite{norrenbrockQPMDiscreteOptimization2025} on their proposed metrics, using the default metrics for compact interpretable models, as well as introduce two metrics, \textit{Plausibility} and \textit{Patch Contextualisation}.

We desire our representations to be diverse \cite{jakkola_diverse}, contrastive \cite{Lipton_1990}, general and compact \cite{readExplanatoryCoherenceSocial1993} as such representations are ideally suited for generating human-interpretable explanations \cite{MILLER20191}. In order to quantify this, we utilise interpretability measures already introduced in \citet{norrenbrockQPMDiscreteOptimization2025}. We employ SID@5, which considers the spatial distinctiveness of feature maps to quantify diversity; meaning that high diversity implies feature maps activate diversely, at different spatial positions. Contrastiveness evaluates the overlap \cite{inmanOverlappingCoefficientMeasure1989} between two components of a Gaussian Mixture Model (GMM) fit to the feature distribution; a feature is considered maximally contrastive if represented by two entirely non-overlapping distributions. Lastly, we evaluate Class-Independence, which measures the proportion of zero-based feature activations across the dataset that is not focused on the most relevant class. We use this as a proxy for feature generality, where high class-independence indicates that features capture broad, general concepts rather than being highly specific to a single class. The concrete definitions of the metrics are given in \cref{app:aux_metrics}. \\
To evaluate the model's ability to localise relevant features within the object of interest, we introduce the Plausibility metric, following the terminology of \citet{jacoviFaithfullyInterpretableNLP2020}. This metric quantifies the fraction of the cumulative feature map activation, weighted by feature relevance, that falls within the ground-truth object boundaries. For the CUB-2011 dataset, we utilise the provided segmentation masks \citep{farrell_2022} $\boldsymbol{M} \in \{0,1\}^{N_p}$ as a tokenised form to define these regions. 
Inspired by the poiting game \cite{zhang2016top} and following Grad-CAM by \citet{selvarajuGradCAMVisualExplanations2020}, the feature maps $\boldsymbol{F}$ are weighted by their decisional relevance. %
First, the aggregate weighted feature map $\widetilde{\boldsymbol{F}}$ is calculated as:
\begin{equation}
    \widetilde{\boldsymbol{F}} = \sum_{d \in \mathcal{F}} W_{\hat{c}d} \, \boldsymbol{F}_{d}
\end{equation}
where $\hat{c}$ is the predicted class, and each feature map $\boldsymbol{F}_{d}$ is scaled by its relevance weight $W_{\hat{c}d}$. We define Plausibility as the proportion of the GradCAM map $\widetilde{\boldsymbol{F}}$ activating within the region of interest:
\noindent\begin{minipage}{.6\linewidth}
\begin{equation}
    \text{Plausibility} = \frac{\displaystyle\sum_{p \in \mathcal{P}} \widetilde{F}_{p} \cdot M^{\text{dil}}_p} {\displaystyle\sum_{p \in \mathcal{P}} \widetilde{F}_{p}} 
\end{equation}
\end{minipage}%
\begin{minipage}{.4\linewidth}
\begin{equation}
    \boldsymbol{M}^{\text{dil}} = \boldsymbol{M} \oplus \mathbbm{1}_{3,3}
    \vphantom{\frac{\displaystyle\sum_{p \in \mathcal{P}}}{\displaystyle\sum_{p \in \mathcal{P}}}}
\end{equation}
\end{minipage}
where $p \in \mathcal{P} = \{1, \ldots, N_p\}$ indexes the set of all image patches.
$\boldsymbol{M}^{\text{dil}}$ is generated by dilating the original mask with a $3 \times 3$ identity structuring element $\mathbbm{1}_{3,3}$. This dilation provides a spatial margin for features that might otherwise overlap with the background due to the patch structure, ensuring that activations accurately capturing the object's silhouette are not unfairly penalised by the metric as visualised in \cref{fig:o_mean}. 
In order to obtain a dataset-wide representative value, the mean Plausibility is computed across all images $\boldsymbol{X} \in \mathcal{X}_{\text{train}}$. A higher Plausibility indicates the Grad-CAM visualisation matches human expectations by concentrating relevant features within the region of interest.

To quantify the degree to which the collective patch representations align with the global semantic information, we introduce the \textit{Patch Contextualisation} metric. We first define the mean patch embedding $\overline{\boldsymbol{F}^{\text{froz}}}$ and then calculate its cosine similarity with the global class token $\boldsymbol{f}^{\text{froz}}_{\text{CLS}}$:
\begin{gather}
    \overline{\boldsymbol{F}^{\text{froz}}} = \frac{1}{N_p} \sum_{p \in \mathcal{P}} \boldsymbol{F}^{\text{froz}}_p \\
    \text{Patch Contextualisation} = \cos\left( \overline{\boldsymbol{F}^{\text{froz}}}, \, \boldsymbol{f}^{\text{froz}}_{\text{CLS}} \right)
\end{gather}
A higher score indicates that the average spatial representation is strongly aligned with $\boldsymbol{f}^{\text{froz}}_{\text{CLS}}$, suggesting that the patch embeddings generally share the same semantic direction as the global context. Conversely, a lower score implies a divergence, indicating that the patch representations contain less global information about the image. We utilise this metric to understand differences between different backbones and bring insights into how various DINO models differ. 
\subsection{Main Results}
\label{subsec:main_results}
\newcommand{\greenmark}{\textcolor{green}{\ding{51}}}
\newcommand{\redmark}{\textcolor{red}{\ding{55}}}
    
\begin{table*}[t]
    \centering
    \setlength{\tabcolsep}{3pt}
    \renewcommand{\arraystretch}{1.1}
    \resizebox{0.85\textwidth}{!}{%
    \begin{tabular}{l c cc c cc cc cc}
        \toprule
        \multirow{2}{*}{Method} & \multirow{2}{*}{\begin{tabular}{c} Local. \\ Features \end{tabular}} & \multicolumn{2}{c}{Accuracy $\uparrow$} & \multirow{2}{*}{Plausibil. $\uparrow$} & \multicolumn{2}{c}{SID@5 $\uparrow$} & \multicolumn{2}{c}{Class-Indep. $\uparrow$} & \multicolumn{2}{c}{Contrast. $\uparrow$} \\
        \cmidrule(lr){3-4} \cmidrule(lr){6-7} \cmidrule(lr){8-9} \cmidrule(lr){10-11}
         & & CUB & CARS & & CUB & CARS & CUB & CARS & CUB & CARS \\
        \midrule
        DINOv2 $\boldsymbol{f}_{\text{CLS}}^{\text{froz}}$ Linear Probe & \redmark & $\underline{87.9}$ & $91.7$ & $42.6$ & $50.9$ & $51.5$ & $\textbf{99.2}$ & $\textbf{99.1}$ & $59.2$ & $60.9$ \\
        Dense $\boldsymbol{F}^{\text{froz}}$ & \greenmark & $78.1$ & $92.9$ & $32.7$ & $\textbf{91.8}$ & $\textbf{93.1}$ & $\underline{98.8}$ & $\underline{98.7}$ & $84.5$ & $82.8$ \\
        Resnet50 Baseline \cite{norrenbrockQPMDiscreteOptimization2025} & \greenmark & $83.9$ & $92.5$ & $60.7$ & $57.1$ & $51.5$ & $98.0$ & $97.9$ & $74.6$ & $75.1$ \\
        \midrule
        Resnet50 QPM \cite{norrenbrockQPMDiscreteOptimization2025} & \greenmark & $82.9$ & $92.1$ & $82.9$ & $89.6$ & $88.2$ & $96.8$ & $97.8$ & $93.6$ & $\underline{97.1}$ \\
        DINO-SLDD & \greenmark & $84.6$ & $92.9$ & $78.0$ & $88.7$ & $90.9$ & $94.4$ & $93.9$ & $93.0$ & $94.9$ \\
        DINO-QSENN & \greenmark & $85.4$ & $\underline{93.3}$ & $86.0$ & $\underline{91.5}$ & $\underline{92.6}$ & $93.6$ & $94.0$ & $\underline{94.4}$ & $94.9$ \\
        \midrule
        \midrule
        DINO-QPM (Ours) & \greenmark & $\textbf{88.3}$ & $\textbf{94.0}$ & $\textbf{95.0}$ & $90.1$ & $91.7$ & $93.7$ & $93.7$ & $\textbf{100.0}$ & $\textbf{100.0}$ \\
        DINO-QPM Compact (Ours) & \greenmark & $\textbf{88.3}$ & $\textbf{94.0}$ & $\underline{94.4}$ & $-$ & $-$ & $93.8$ & $93.6$ & $\textbf{100.0}$ & $\textbf{100.0}$ \\
        \bottomrule
    \end{tabular}%
    }
    \caption{Comparison with state-of-the-art interpretable models. We report Accuracy, Plausibility, SID@5, Class-Independence, and Contrastiveness (all metrics in $\%$). Features of a model are localised if they have a direct connection to the feature vector used for classification. The Plausibility metric is evaluated only on CUB-2011 due to the availability of segmentation masks. Dense $\boldsymbol{F}^{\text{froz}}$ is the dense model of DINO-QPM and DINOv2 $\boldsymbol{f}_{\text{CLS}}^{\text{froz}}$ Linear Probe is a linear probe \cite{chenSimpleFrameworkContrastive2020} trained on top of the frozen \texttt{CLS} representation. For DINO-SLDD and DINO-QSENN, we employ a pipeline closely resembling the one described in \cref{sec:method}, with the exception of the feature selection mechanisms, which follow \citet{norrenbrockTake5Interpretable2023} and \citet{norrenbrockQSENNQuantizedSelfExplaining2024}, respectively.}
    \label{tab:comparison_sota}
\end{table*}

\begin{table}[t]
    \centering
    \small
    \setlength{\tabcolsep}{4pt}
    \renewcommand{\arraystretch}{1.1}
    \resizebox{\columnwidth}{!}{%
    \begin{tabular}{l c c c}
        \toprule
        Method & \begin{tabular}[b]{@{}c@{}} $\#$ Features \\ ($N_f$) \end{tabular} & \begin{tabular}[b]{@{}c@{}} $\#$ Features \\ per class ($N_f^c$) \end{tabular} & \begin{tabular}[b]{@{}c@{}} Avg. Training \\ Time per Epoch [s] \end{tabular} \\
        \midrule
        DINOv2 $\boldsymbol{f}_{\text{CLS}}^{\text{froz}}$ Linear Probe & $768$ & $768$ & $3$ \\
        DINOv2 $\boldsymbol{F}^{\text{froz}}$ Baseline & $512$ & $512$ & $6$ \\
        Resnet50 Baseline & $2048$ & $2048$ & $40$ \\
        \midrule
        Resnet50 QPM & $50$ & $5$ & $40$ \\
        DINO-SLDD & $50$ & $5$ & $6$ \\
        DINO-QSENN & $50$ & $5$ & $6$ \\
        \midrule
        \midrule
        DINO-QPM (Ours) & $50$ & $5$ & $6$ \\
        DINO-QPM Compact (Ours) & $40$ & $4$ & $6$ \\
        \bottomrule
    \end{tabular}%
    }
    \caption{Comparison of model complexity and training efficiency. We report the total number of features ($N_f$), features per class ($N_f^c$), and the average training time per epoch in seconds on the referenced hardware (\cref{app:implementation_details}).}
    \label{tab:features_overview}
\end{table}

In this section, we present our main experimental results. In \cref{tab:comparison_sota}, we evaluate the interpretability metrics introduced in \cref{subsec:metrics} alongside the accuracy of each approach, while \cref{tab:features_overview} outlines compactness and training time. Across both datasets, DINO-QPM achieves the highest overall accuracy and outperforms the baselines and other approaches in plausibility, demonstrating a substantial increase in both metrics compared to dense representations. Furthermore, DINO-QPM exhibits strong interpretability, reaching high feature diversity (SID@5) and Class-Independence while securing the highest Contrastiveness. Crucially, it accomplishes this while extracting features that are localised by design and maintaining representational compactness, aligning with other interpretability-enhancing approaches (or even surpassing them, as seen with DINO-QPM Compact), but contrasting sharply with the dense baselines. From a computational perspective, relying on a frozen DINOv2 backbone drastically reduces the required training resources, particularly when compared to the ResNet-50 QPM, as it enables pre-computation of patch and \texttt{CLS} token embeddings. Notably, our linear probe reaches lower accuracy than reported in the original paper \cite{oquabDINOv2LearningRobust2024}. This discrepancy arises because they use a different evaluation scheme, which includes output of up to four layers of the ViT while allowing the concatenation of average-pooled patch embeddings to the \texttt{CLS} tokens.
The supplementary material provides full results with standard deviations (\cref{app:detailed_results}) and further visualisations (\cref{app:visualisations}) contrasting DINO-QPM's faithful explanations against the baseline's ungrounded attributions across successes, baseline failures, and interpretable failure cases.

\subsection{Ablation Studies}
\label{subsec:ablations}
In this section, we conduct a series of ablation studies to systematically evaluate the architectural and hyperparameter choices for DINO-QPM which influence predictive performance and model interpretability.

\begin{table}[t]
    \centering
    \small
    \setlength{\tabcolsep}{4pt}
    \renewcommand{\arraystretch}{1.1}
    \begin{tabular}{c c c c c}
        \toprule
        \begin{tabular}[b]{@{}c@{}} Input \\ Source \end{tabular} & \begin{tabular}[b]{@{}c@{}} Loc. \\ Feat. \end{tabular} & Registers & Acc. $\uparrow$ & Plausibil. $\uparrow$ \\
        \midrule
        \multirow{2}{*}{$\boldsymbol{f}_{\text{CLS}}^{\text{froz}}$} & \multirow{2}{*}{\redmark} & \redmark & $87.3$ & $\underline{96.9}$ \\
         & & \greenmark & $\underline{87.6}$ & $\mathbf{99.2}$ \\
        \midrule
        \multirow{2}{*}{$\boldsymbol{F}^{\text{froz}}$} & \multirow{2}{*}{\greenmark} & \redmark & $83.3$ & $73.5$ \\
         & & \greenmark & $\mathbf{88.3}$ & $95.0$ \\
        \bottomrule
    \end{tabular}
    \caption{Ablation study showing the impact of input source and register tokens on CUB-2011. The local features column indicates whether the input representation preserves spatial information.}
    \label{tab:registers_ablation}
\end{table}

To compare DINO-QPM with a variant utilising the CLS token, we define its predictive pipeline as follows: $\hat{c}=\arg\max\big\{\text{BLDD}(\text{MLP}(\boldsymbol{f}^{\text{froz}}_{\text{CLS}}))\big\}$. To maintain a feature map equivalent, we apply the same MLP to the frozen patch embeddings and utilise the output as our feature maps. 

In \Cref{tab:registers_ablation}, we show that when using registers, achieving high accuracy only requires patch embeddings ($\boldsymbol{F}^{\text{froz}}$), allowing us to discard the CLS representation ($\boldsymbol{f}^{\text{froz}}_{\text{CLS}}$). Despite $\boldsymbol{f}^{\text{froz}}_{\text{CLS}}$ scoring high on Plausibility, it lacks a direct connection between the features and their maps (i.e., no localised features). Therefore, high Plausibility is less meaningful than for DINO-QPM ($\boldsymbol{F}^{\text{froz}}$), which possesses this property. %

\begin{table}[t]
    \centering
    \resizebox{0.85\columnwidth}{!}{%
    \begin{tabular}{@{}llcc@{}}
        \toprule
        \multirow{2}{*}{Backbone Architecture} & \multirow{2}{*}{Acc. $\uparrow$}  & \multirow{2}{*}{Patch Context. $\uparrow$} \\
         & & & \\ %
        \midrule
        DINO ViT-B/16  & $37.1$ & $8.9$ \\
        \addlinespace[0.3em] 
        DINOv2 ViT-S/14 Reg. & $83.4$ & $42.9$ \\
        DINOv2 ViT-B/14 Reg.  & $88.3$ & $43.9$ \\
        DINOv2 ViT-L/14 Reg. & $86.5$ & $2.2$ \\
        \bottomrule
    \end{tabular}%
    }
    \caption{Comparison of different backbone architectures on CUB-2011. We compare DINO with DINOv2 and different ViT sizes and report accuracy alongside Patch Contextualisation.}
    \label{tab:dino_comparison}
\end{table}

Comparing different backbone sizes for DINOv2 (\cref{tab:dino_comparison}), we observe our proposed approach can be applied to all of them. However, the accuracy declines with larger and smaller backbones. For ViT-L, this might be due to low Patch Contextualisation. The model has insufficient global context in its patch embeddings, which hinders higher accuracy. For ViT-S we tested DINOv2 $\boldsymbol{f}_{\text{CLS}}^{\text{froz}}$ Linear Probe which reaches similar accuracy. Therefore, we conclude that the disproportionate (compared to \cite{oquabDINOv2LearningRobust2024}) decline is not attributable to DINO-QPM but their use of a different evaluation scheme (see \cref{subsec:main_results}). Hence, we demonstrate a previously unexplored difference between DINOv2 backbone sizes that affects interpretable downstream performance and may guide further research. %
Furthermore, DINO-QPM exhibits poor accuracy on DINO ViT-B/16 combined with low Patch Contextualisation. Frozen DINO in general does not perform well on CUB-2011 when evaluating a linear probe on  $\boldsymbol{f}_{\text{CLS}}^{\text{froz}}$ \cite{liuDataLanguageSupervision2025, marksCloserLookBenchmarking2024} and the low amount of global image information in the patch embeddings exacerbates this.  %
\begin{figure}[t]
    \centering
    \includegraphics[width=0.82\columnwidth]{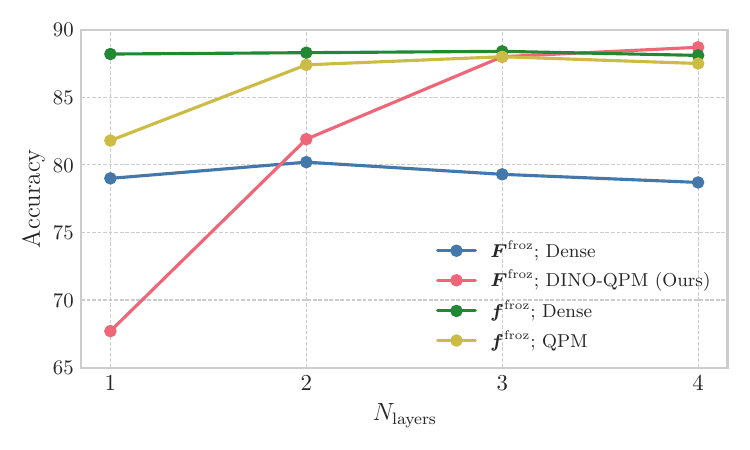} %
    \caption{Impact of the number of MLP layers ($N_{\text{layers}}$) on classification accuracy. The plot compares the accurcy on CUB-2011 of using frozen patch-level feature maps ($\boldsymbol{F}^{\text{froz}}$) versus the global feature vector ($\boldsymbol{f}_{\text{CLS}}^{\text{froz}}$) for both Dense and QPM}
    \label{fig:ablation_n_layers}
\end{figure}

While the number of layers in the MLP has no measurable impact on the dense model, when introducing sparsity, especially using the patch embeddings, we observe a huge increase in model accuracy (\cref{fig:ablation_n_layers}). Notably, comparing dense and QPM accuracy for higher $N_{\text{layers}}$, the QPM is able to outperform its dense model by almost $10 \%$ using $\boldsymbol{F}^{\text{froz}}$.

\begin{figure}[t]
    \centering
    \begin{subfigure}[b]{0.49\columnwidth}
        \includegraphics[width=\linewidth]{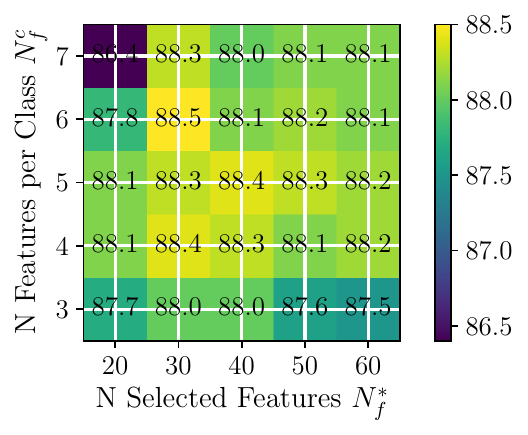}
        \caption{CUB2011}
        \label{fig:heatmap_cub}
    \end{subfigure}
    \hfill
    \begin{subfigure}[b]{0.49\columnwidth}
        \includegraphics[width=\linewidth]{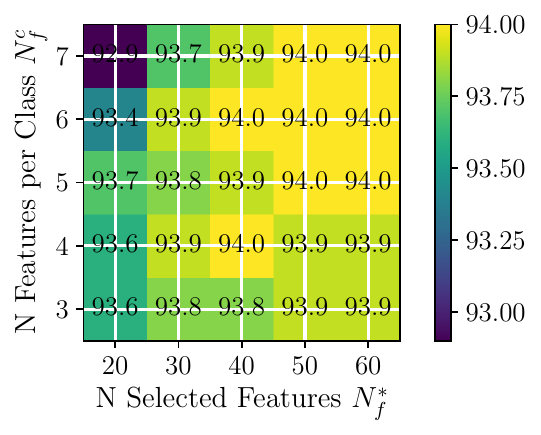}
        \caption{Stanford Cars}
        \label{fig:heatmap_cars}
    \end{subfigure}
    \caption{Impact of sparsity constraints on classification accuracy. We vary the total number of selected features $N_f^*$ and the number of features assigned per class $N_f^c$. Compared to the QPM \cite{norrenbrockQPMDiscreteOptimization2025} we observe relatively low impact on accuracy. }
    \label{fig:hyperparam_heatmaps}
\end{figure}

We evaluate the compactness of DINO-QPM across Stanford Cars and CUB-2011 w.r.t the total number of selected features $N_f^*$ and the features per class $N_f^c$ (\cref{fig:hyperparam_heatmaps}).
We observe that the overall variation in accuracy across both datasets is remarkably low ($<1\%$). This contrasts with the higher sensitivity reported in \citet{norrenbrockQPMDiscreteOptimization2025}. Hence, Dino-QPM enables higher accuracy at a higher compactness, e.g. using our Compact Model in \cref{tab:comparison_sota}.

\label{subsec:l1_fm}
\begin{figure}[t]
    \centering
    \includegraphics[width=0.85\columnwidth]{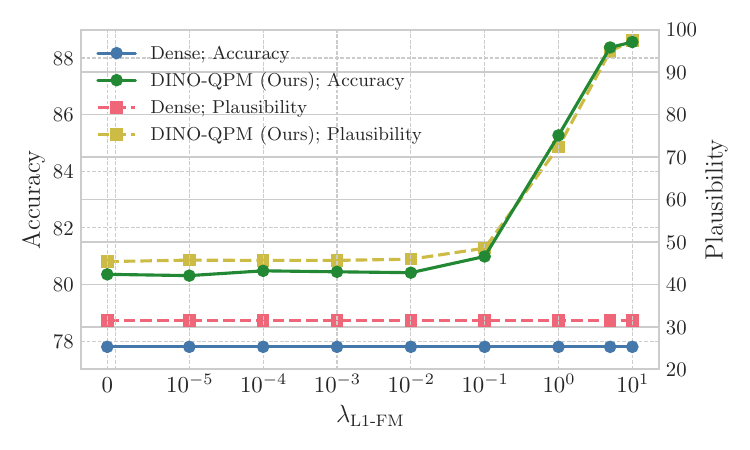}%
    \caption{Effect of varying the $\mathcal{L}_{\text{L1-FM}}$ weight during finetuning compared to the dense model. Higher penalty weights yield substantial improvements in both accuracy and plausibility metrics.}
    \label{fig:l1_fm}
\end{figure}

\begin{figure}[t]
    \centering
    \includegraphics[width=0.48\textwidth]{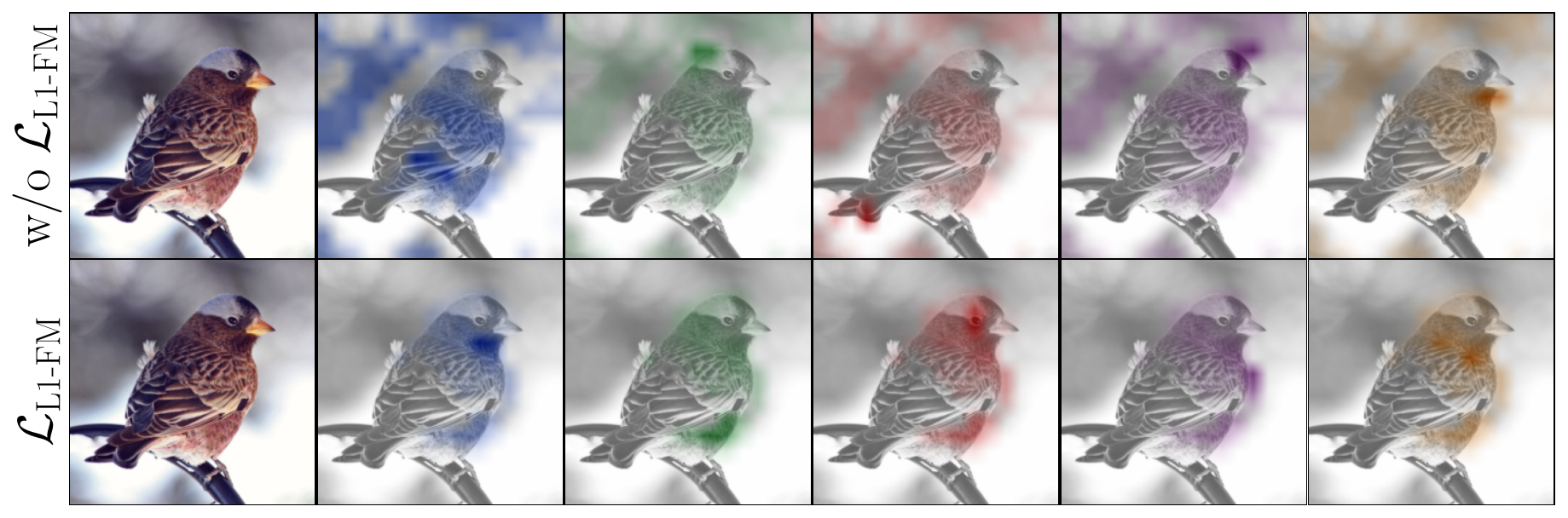}
    \caption{Qualitative ablation of $\mathcal{L}_{\text{L1-FM}}$ on the Gray-crowned Rosy Finch. Without $\mathcal{L}_{\text{L1-FM}}$ (top row), feature activations exhibit background noise and spatial scatter. Adding $\mathcal{L}_{\text{L1-FM}}$ (bottom row) suppresses this noise, resulting in distinct activations semantically localised to specific object parts.}
    \label{fig:no_l1_fm}
\end{figure}

A significant increase in model accuracy which highly correlates with Plausibility is observed when increasing $\lambda_{\text{L1-FM}}$ (\cref{fig:l1_fm}), the weight of the L1 sparsity loss on the feature maps. A potential explanation is that the regularisation forces the model to concentrate its activation mass on object regions relevant for classification which then has a tremendously positive effect on accuracy (\cref{fig:no_l1_fm}).

%% file: sec/6_conclusion.tex
\section{Conclusion}
\label{sec:conclusion}
In this work, we introduced DINO-QPM as a compactness-based interpretability adapter applied on top of frozen backbones like DINOv2 \cite{oquabDINOv2LearningRobust2024} to achieve globally interpretable image classification. By establishing a direct connection between the feature vector and its corresponding maps via a non-standard average-pooling approach, DINO-QPM achieves exceptional spatial localisation, substantiating its highly faithful decision process. Paired with inherently contrastive, diverse, and general features, our approach outperforms the uninterpretable DINOv2 linear probe, as well as other baselines and applicable approaches, on fine-grained image classification, delivering state-of-the-art interpretability on top of frozen visual foundation models while maintaining exceptional accuracy.

%% file: sec/7_acknowledgements.tex
\section*{Acknowledgements}
Financial support for this research was provided by the MWK of Lower Saxony through the Hybrint (VWZN4219) and LCIS (VWZN4704) projects. Furthermore, funding was granted by the Deutsche Forschungsgemeinschaft (DFG) as part of Germany’s Excellence Strategy for the PhoenixD (EXC2122) and Quantum Frontiers (EXC2123) Clusters of Excellence, and by the European Union under grant agreement no. 101136006 – XTREME.

%% file: sec/X_suppl.tex
\clearpage
\setcounter{page}{1}
\maketitlesupplementary

\section{Feature Diversity Loss}
\label{app:fdl}
To reduce conceptual ambiguity between features, \citet{norrenbrockTake5Interpretable2023} introduced the Feature Diversity Loss, hereafter referred to as $\mathcal{L}_{\text{div}}$. The objective of $\mathcal{L}_{\text{div}}$ is to encourage the representation of distinct, mutually independent concepts within the features, thereby enhancing the degree of model interpretability.
Let $i \in \mathcal{I} = \{ 1, \dots, W_f \}$ and $j \in \mathcal{J} = \{ 1, \dots, H_f \}$ denote the spatial dimensions of the feature map $\boldsymbol{F}^d$ associated with feature $d \in \mathcal{F} = \{1, \dots, N_f \}$. Furthermore, let $\boldsymbol{W}_{(\hat{c})}$ represent the row of the weight matrix $\boldsymbol{W}$ corresponding to the predicted class $\hat{c}$, and $W_{\hat{c}d}$ represent the specific entry for class $\hat{c}$ and feature $d$. The diversity loss $\mathcal{L}_{\text{div}}$ is defined by the following equations:

\begin{equation}
\label{eq:fdl}
\mathcal{L}_{\text{div}} = -\sum_{i \in \mathcal{I}} \sum_{j \in \mathcal{J}} \max_{d \in \mathcal{F}} \hat{S}_{ij}^d
\end{equation}

where the weighted diversity maps $\hat{S}_{ij}^d$ are computed for all $i \in \mathcal{I}, j \in \mathcal{J}$, and $d \in \mathcal{F}$ according to:

\begin{equation}
\label{eq:s_hat}
\hat{S}_{ij}^d = \frac{\exp(F_{ij}^d)}{\sum\limits_{i' \in \mathcal{I}} \sum\limits_{j' \in \mathcal{J}} \exp(F_{i' j'}^d)} \frac{f_d}{\max\limits_{d' \in \mathcal{F}} f_{d'}} \frac{|W_{\hat{c} d}|}{\|\boldsymbol{W}_{(\hat{c})}\|_2}
\end{equation}

\cref{eq:s_hat} employs the softmax function to normalize $\boldsymbol{F}^d$ across its spatial dimensions $i$ and $j$. Simultaneously, the feature map is weighted in the feature dimension $d$: first, by the value of feature $d$ relative to the maximum of the feature vector, and second, by scaling $W_{\hat{c}d}$ relative to the $L_2$-norm of the weights for all features associated with the predicted class. These components serve to highlight decision-relevant features. \cref{eq:fdl} then ensures that the normalized feature maps $\hat{\boldsymbol{S}}^d$ focus on distinct spatial regions. Overall, $\mathcal{L}_{\text{div}}$ acts as a regularizer to the standard cross-entropy loss, resulting in a total objective function:

\begin{equation}
\mathcal{L}_{\text{total}} = \mathcal{L}_{\text{CE}} + \beta \mathcal{L}_{\text{div}}
\end{equation}

where $\beta \in \mathbb{R}_+$ is a weighting hyperparameter.

\section{Definition of Additional Interpretability Metrics}
\label{app:aux_metrics}
To assess model interpretability, we apply several metrics following \citet{norrenbrockTake5Interpretable2023, norrenbrockQSENNQuantizedSelfExplaining2024, norrenbrockQPMDiscreteOptimization2025}. Since interpretability is multifaceted, multiple metrics addressing distinct concepts are necessary. 

Throughout this section, we utilise the following notation for index sets: $i \in \mathcal{I} = \{1, \dots, W_f\}$ and $j \in \mathcal{J} = \{1, \dots, H_f\}$ denote spatial dimensions, $d \in \mathcal{F} = \{1, \dots, N_f\}$ denotes the feature indices, $c \in \mathcal{C} = \{1, \dots, N_c\}$ denotes class indices, and $x \in \mathcal{X}_{\text{train}}$ represents samples from the training dataset.

\subsection{SID@k}
Similar to the $\mathcal{L}_{\text{div}}$ presented in \cref{app:fdl}, we utilise the Scale-Invariant Diversity (SID) from \citet{norrenbrockQPMDiscreteOptimization2025}. This metric measures the distinctiveness between the feature maps $\boldsymbol{F}^d$ of each feature $d$.

\begin{equation}
\label{eq:normed_fm}
\begin{split}
    \widehat{F}_{ij}^d &= \frac{1}{F_{\text{avg}}^d} \, F_{ij}^d \\
    \text{with} \quad F_{\text{avg}}^d &= \frac{1}{W_f H_f} \sum_{i \in \mathcal{I}}\sum_{j \in \mathcal{J}} \, \left| F_{ij}^d \right|
\end{split}
\end{equation}

First, the feature maps $\boldsymbol{F}^d$ are normalized by their absolute mean $F_{\text{avg}}^d$ for all $d \in \mathcal{F}$ (\cref{eq:normed_fm}).

\begin{equation}
\begin{split}
    \widehat{S}_{ij}^d &= \frac{\exp\left({\widehat{F}_{ij}^d}\right)}{\displaystyle \sum_{i' \in \mathcal{I}}\sum_{j' \in \mathcal{J}} \exp\left({\widehat{F}_{i'j'}^d}\right)} \\
    &\quad \forall \; i \in \mathcal{I}, j \in \mathcal{J}, d \in \mathcal{F}
\end{split}
\end{equation}

A softmax function is then applied to the normalized feature maps $\widehat{\boldsymbol{F}}^{d}$ to obtain $\widehat{\boldsymbol{S}}^d$.

\begin{equation}
\begin{split}
     \hat{S}^{\, \max}_{ij} &= \max\limits_{d \, \in \, \mathcal{F}_k}\,\hat{S}_{ij}^d \\
     &\quad \forall \; i \in \mathcal{I}, j \in \mathcal{J}
\end{split}
\end{equation}

Subsequently, along the feature dimension, the maximum of the $k$ highest-weighted, normalised feature maps $\widehat{\boldsymbol{S}}^d$ is computed for each spatial element. Here, $\mathcal{F}_k \subset \mathcal{F}$ denotes the subset of exactly those $k$ features associated with the highest weights.
The SID@k is defined as the sum over all $\hat{S}^{\, \max}_{ij}$, normalized by $k$.

\begin{equation}
    \textrm{SID@k} = \frac{1}{k} \sum_{i \in \mathcal{I}}\sum_{j \in \mathcal{J}} \, \hat{S}_{ij}^{\, \max} 
\end{equation}

\subsection{Class-Independence $\tau$}
\citet{norrenbrockQPMDiscreteOptimization2025} propose Class-Independence $\tau$ as a measure of whether features represent a general or a class-specific concept.
For this purpose, the individual feature values $f_x^d$ per data point and feature are first normalized over the entire training dataset such that their minimum is $0$.

\begin{equation}
\label{eq:f_min}
\begin{split}
    f_{x, \text{norm}}^{d} &= f_{x}^d - f_{\min}^d \\
    \text{with} \quad f_{\min}^d &= \min_{x' \in \mathcal{X}_{\text{train}}} f_{x'}^d
\end{split}
\end{equation}

The resulting $f_{x, \text{norm}}^{d}$ values (for all $x \in \mathcal{X}_{\text{train}}, d \in \mathcal{F}$) are then used in conjunction with the label vector $\boldsymbol{l}^{c}$—where $l_x^{c} = 1$ if $x$ belongs to class $c$, and $0$ otherwise—to obtain $\varphi^{cd}$. This term indicates how strongly feature $d$ focuses on class $c$.

\begin{equation}
\begin{split}
    \varphi^{cd} &= \frac{\displaystyle\sum_{x \in \mathcal{X}_{\text{train}}} l_x^{c} \cdot f_{x,\text{norm}}^{d}}{\displaystyle\sum_{x \in \mathcal{X}_{\text{train}}} f_{x,\text{norm}}^{d}} \\
    &\quad \forall \; c \in \mathcal{C}, d \in \mathcal{F}
\end{split}
\end{equation}

By selecting the class $c$ on which each feature $d$ focuses most strongly and averaging these values, the Class-Dependence is obtained. The Class-Independence $\tau$ is then defined as the complement of the Class-Dependence relative to 1.

\begin{equation}
    \tau = 1 - \frac{1}{N_f} \sum_{d \, \in \, \mathcal{F}} \max_{c \, \in \, \mathcal{C}} \, \varphi^{cd}
\end{equation}

\subsection{Contrastiveness}
Let the empirical feature distribution $\hat{p}(f_x^d)$ be a normalized histogram over the vector $\boldsymbol{f}^d$, containing the feature values of a feature $d$ for all training data.
To measure contrastiveness, a Gaussian Mixture Model (GMM) with two components is constructed for each feature distribution $\hat{p}(f_x^d)$, yielding two normal distributions $\mathcal{N}_1^d$ and $\mathcal{N}_2^d$. The first component models the non-activation region, while the second approximates the activation region.

\begin{equation}
\begin{split}
    \textrm{Contrastiveness} &= 1 - \frac{1}{N_f} \sum_{d \, \in \, \mathcal{F}} \textrm{Overlap}\left( \mathcal{N}_1^d, \mathcal{N}_2^d \right)
\end{split}
\end{equation}

Contrastiveness results as the expected non-overlap \cite{inmanOverlappingCoefficientMeasure1989} of the two distributions $\mathcal{N}_1^d$ and $\mathcal{N}_2^d$. Thus, a feature is considered (maximally) contrastive if and only if it can be represented by a bimodal distribution of two non-overlapping distribution functions.

\section{Implementation Details}
\label{app:implementation_details}
All input images are resized to $224 \times 224$ pixels and normalised according to the dataset mean values. 

Unless otherwise specified, the Multi-Layer Perceptron (MLP) consists of four layers featuring ReLU activation and batch normalisation. The number of features is set to $N_f=512$, and the number of neurons in the hidden layers is $N_{\text{hidden}}=2048$. 
To manage the learning rate, a schedule-free approach following \citet{defazioRoadLessScheduled2024} is employed in combination with Adam as our optimiser. \\
In dense training we train for $40$ epochs using a weight decay of $7 \cdot 10^{-4}$ with batch size $32$ and a start learning rate of $10^{-3}$. In our BLDD layer we use a dropout of $0.2$. Besides $\mathcal{L}_{\text{CE}}$ we use $\mathcal{L}_{\text{div}}$ with $\lambda_{\text{div}}=0.5$. \\
In fine-tuning training we train for $50$ epochs with a weight decay of $8 \cdot 10^{-4}$ with batch size $32$ and a start learning rate of $5 \cdot 10^{-3}$. We explicitly do not use dropout in our BLDD layer when fine-tuning. Besides $\mathcal{L}_{\text{CE}}$ we use $\mathcal{L}_{\text{div}}$ with $\lambda_{\text{div}}=1$, $\mathcal{L}_{\text{L1-FM}}$ with $\lambda_{\text{L1-FM}}=5$ and $\mathcal{L}_{\text{L1-FV}}$ with $\lambda_{\text{L1-FV}}=1$. \\
The QP is solved using Gurobi \cite{gurobioptimizationllcGurobiOptimizerReference2024}, while the neural network architectures are implemented in PyTorch \cite{paszkePyTorchImperativeStyle2019}. 
For measuring the training time in \cref{tab:features_overview}, we used an NVIDIA GeForce RTX 3090 GPU combined with an 11th Gen Intel(R) Core(TM) i9-11900K @ 3.50GHz CPU. 

We report the mean and standard deviation across 5 random seeds for all models, with the exception of our DINO-QPM base configuration, which was evaluated over 15 seeds due to a configuration oversight. This larger sample size provides a more precise estimate of the mean without introducing any bias into the comparison.

\section{Impact of Auxiliary Losses}
\label{app:losses}
\begin{figure*}[t]
    \centering
    \includegraphics[width=\columnwidth]{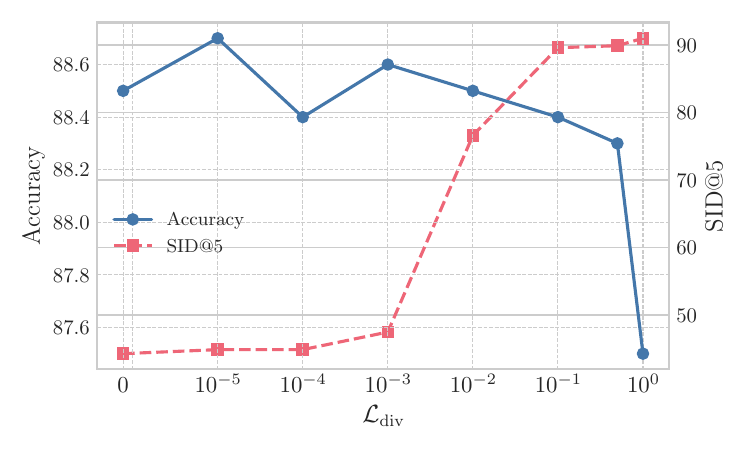}%
    \caption{Accuracy and Feature Diversity (SID@5) on CUB-2011 across variations of the $\mathcal{L}_{\text{div}}$ weight during dense and finetuning training.}
    \label{fig:fdl}
\end{figure*}

The $\mathcal{L}_{\text{div}}$ loss, as proposed by \citet{norrenbrockTake5Interpretable2023} and introduced in detail in \cref{app:fdl}, is analysed here. \cref{fig:fdl} illustrates the influence of $\mathcal{L}_{\text{div}}$ on accuracy and SID@5. Notably, increasing the weight of this loss has a strong positive correlation with SID@5. Hence, the lightweight interpretability adapter can be steered similarly to the end-to-end trained models. %

\begin{figure*}
    \centering
    \includegraphics[width=\columnwidth]{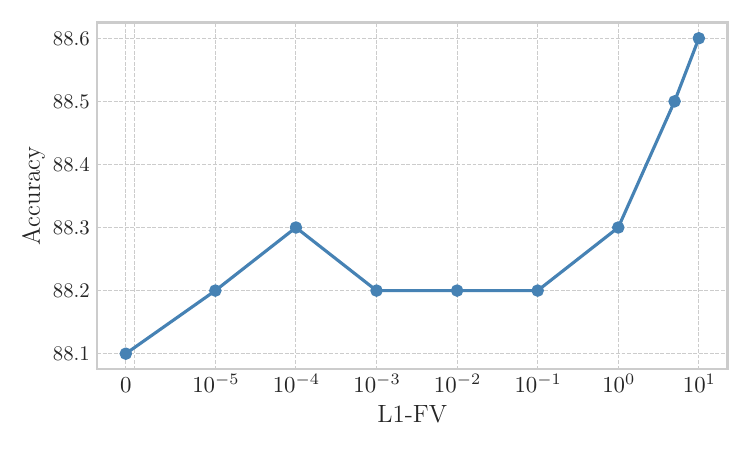}%
    \caption{Impact of the $\mathcal{L}_{\text{L1-FV}}$ on CUB-2011 accuracy during finetuning. }
    \label{fig:l1_fv}
\end{figure*}

In the finetuning stage, besides the aforementioned $\mathcal{L}_{\text{L1-FM}}$ an additional L1 regularization loss $\mathcal{L}_{\text{L1-FV}}$ on the feature vector is introduced alongside $\mathcal{L}_{\text{div}}$. Looking at \cref{fig:l1_fv} we observe its positive impact on accuracy.

\section{Impact of MLP Depth}
\begin{figure*}[t]
    \centering
    \includegraphics[width=\columnwidth]{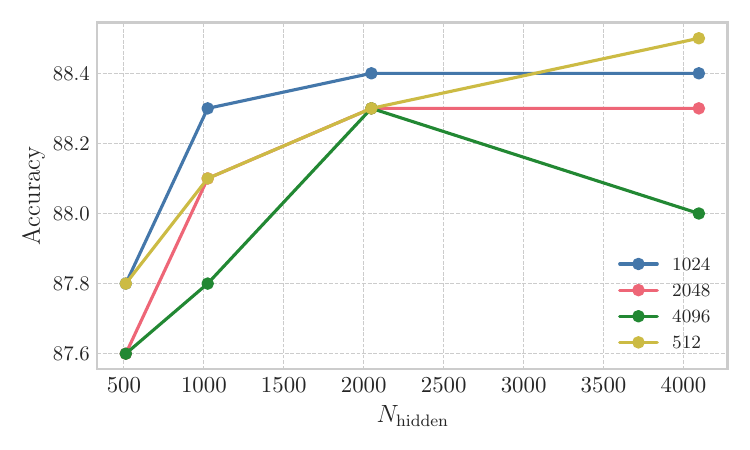}
    \caption{Mean finetuning accuracy on CUB-2011 for various numbers of features $N_f$ across a range of hidden layer neurons $N_{\text{hidden}}$ in the MLP. We observe small accuracy gains up until $N_{\text{hidden}}=2048$ regardless of the number of features $N_f$.} 
    \label{fig:n_f-n_hidden}
\end{figure*}

\cref{fig:n_f-n_hidden} illustrates the accuracy plotted against the number of neurons in the MLP's hidden layers $N_{\text{hidden}}$. Small accuracy gains are observed up to $N_{\text{hidden}}=2048$, regardless of the number of features $N_f$ which is why we chose $N_{\text{hidden}}=2048$ and $N_f=512$, obtaining optimal accuracy while minimising compactness.

\clearpage %
\onecolumn %

\section{Visualisations}
\label{app:visualisations}
\subsection{Class Comparisons}
\begin{figure}[!ht]
    \centering
    \includegraphics[width=\textwidth]{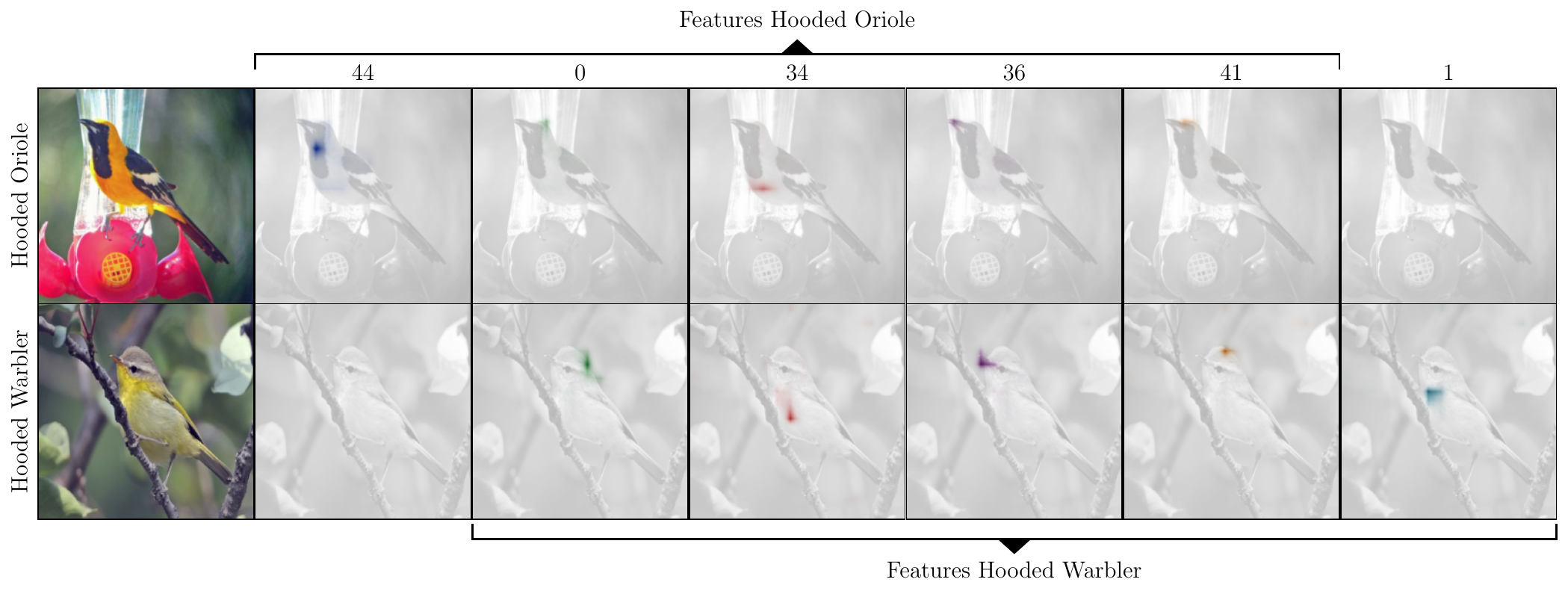}
    \caption{Faithful global interpretability on CUB-2011: DINO-QPM autonomously discovers the 5 diverse, generalisable features for each class used to represent the Hooded Oriole and Hooded Warbler, completely without external supervision. The probed QPM distinguishes them using their evidently different throat. }
    \label{fig:class_comp1}
\end{figure}
\begin{figure}[!ht]
    \centering
    \includegraphics[width=\textwidth]{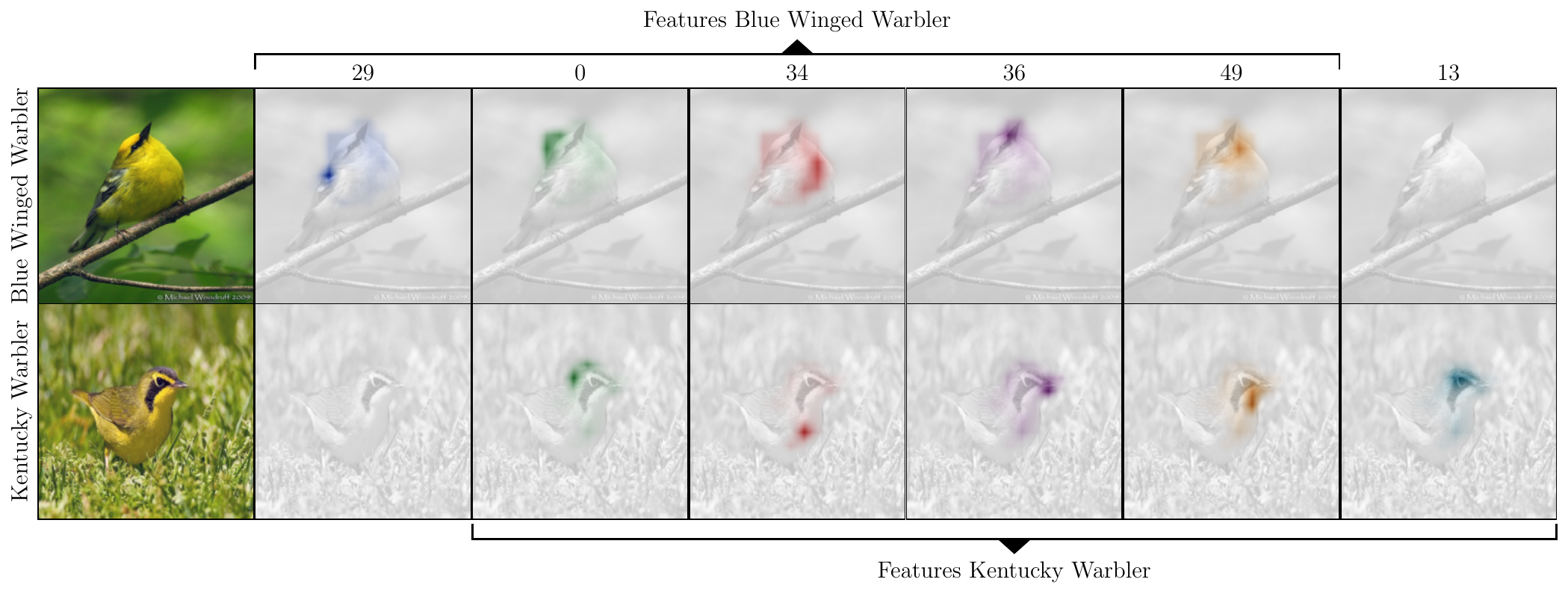}
    \caption{Faithful global interpretability on CUB-2011: DINO-QPM autonomously discovers the 5 diverse, generalisable features for each class used to represent the Blue Winged Warbler and Kentucky Warbler, completely without external supervision. The probed QPM distinguishes them using their evidently different eye area. }
    \label{fig:class_comp2}
\end{figure}
\begin{figure}[!ht]
    \centering
    \includegraphics[width=\textwidth]{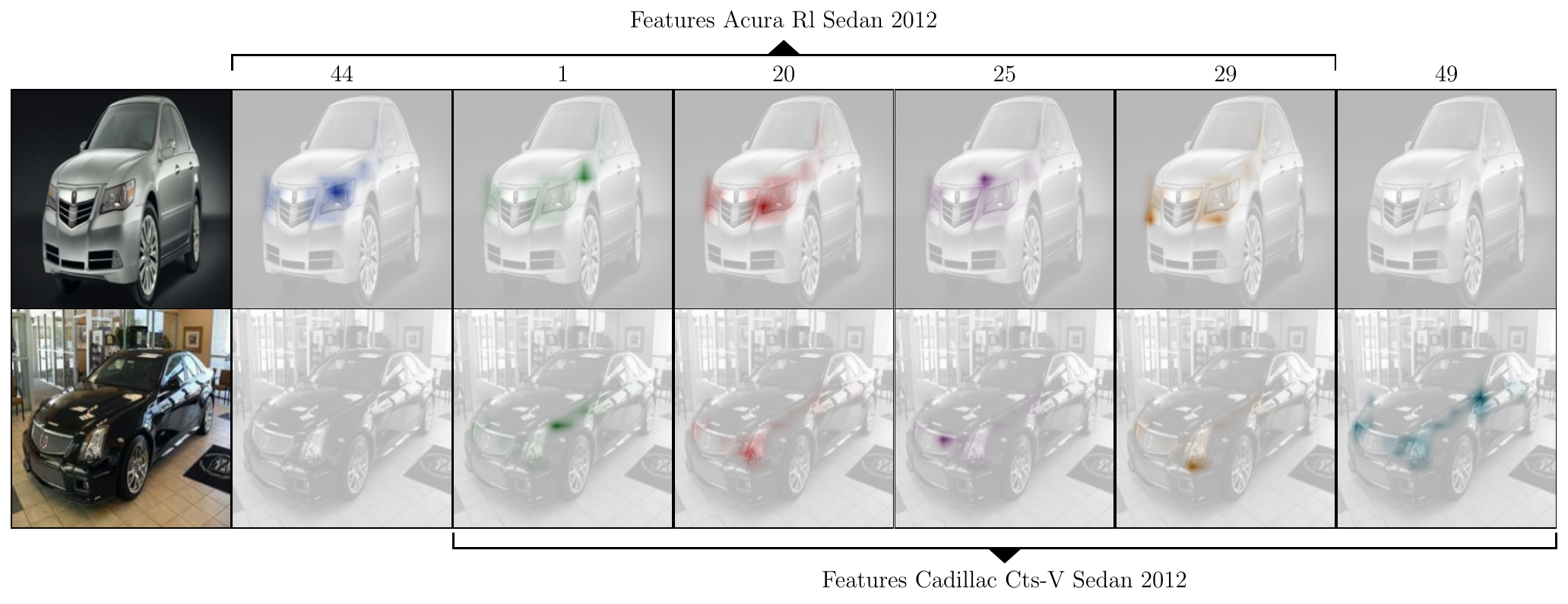}
    \caption{Faithful global interpretability on Stanford Cars: DINO-QPM autonomously discovers the 5 diverse, generalisable features for each class used to represent the Acura Rl Sedan 2012 and Cadillac Cts-V Sedan 2012, completely without external supervision. The probed QPM distinguishes them using their evidently different headlights. }
    \label{fig:class_comp3}
\end{figure}
\begin{figure}[!ht]
    \centering
    \includegraphics[width=\textwidth]{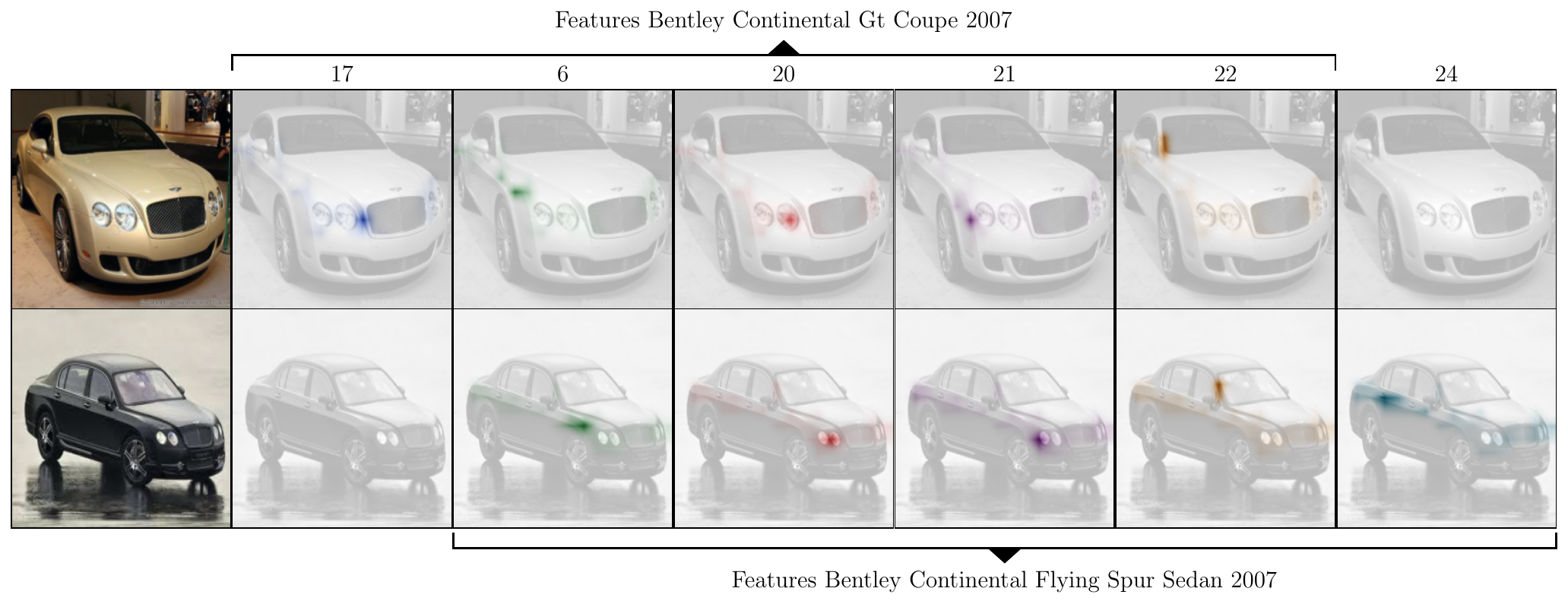}
    \caption{Faithful global interpretability on Stanford Cars: DINO-QPM autonomously discovers the 5 diverse, generalisable features for each class used to represent the Bentley Continental Gt Coupe 2007 and Bentley Continental Flying Spur Sedan 2007, completely without external supervision. The probed QPM distinguishes them using their evidently different door configurations. The probed QPM distinguishes them using their evidently different door configurations. As the most prominent distinguishing factor is the number of doors, the model's non-overlapping features for the Flying Spur (Sedan) specifically highlight the rear doors and rear door handles, which the GT (Coupe) lacks}
    \label{fig:class_comp4}
\end{figure}

\clearpage
\newpage

\subsection{Dense \texorpdfstring{$\boldsymbol{F}^{\text{froz}}$}{F\textasciicircum{froz}} Failure vs. DINO-QPM Correct Classification}
\begin{figure}[!ht]
    \centering
    \includegraphics[width=0.9\textwidth]{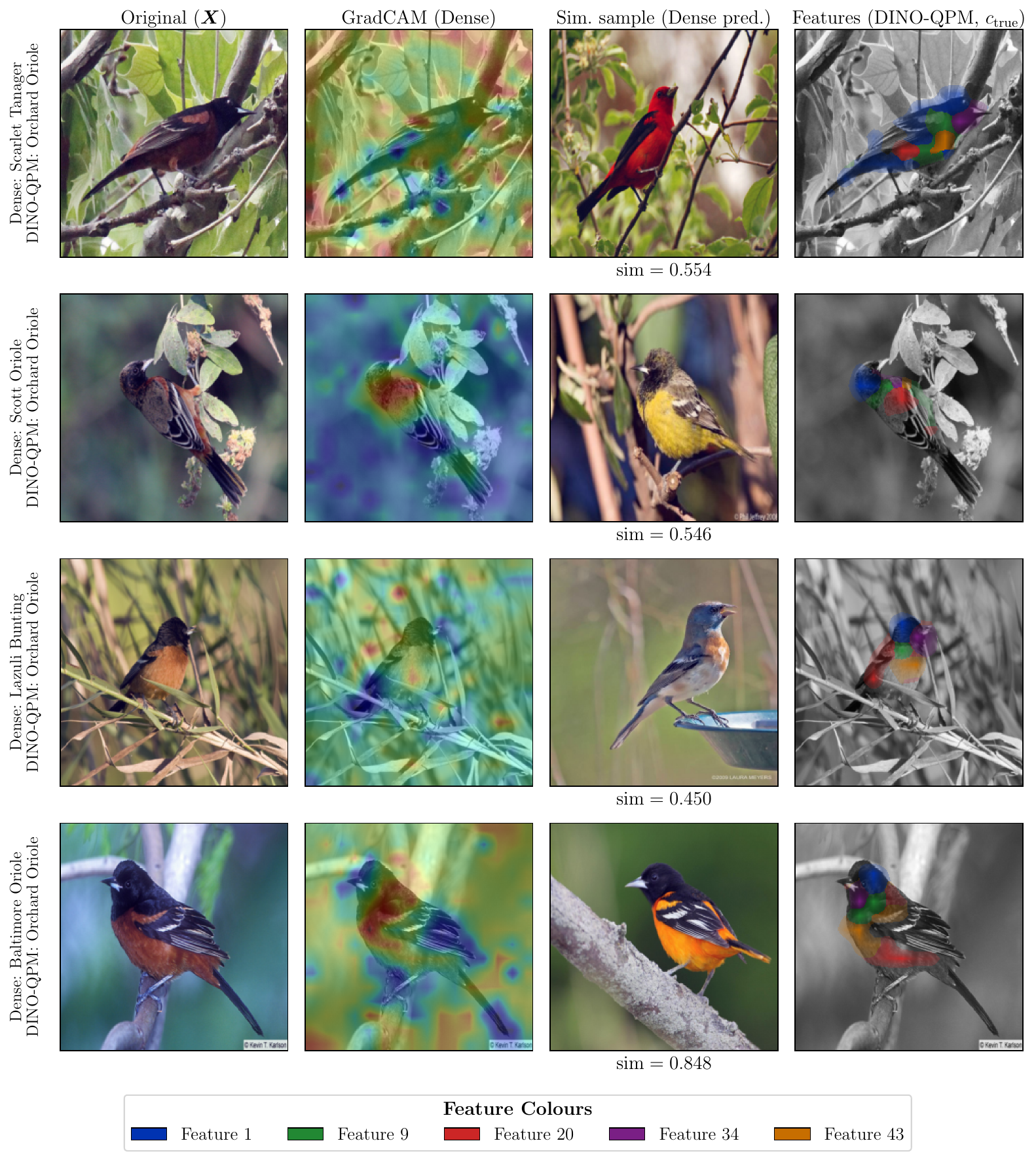}
    \caption{Comparison on the \textit{Orchard Oriole} (CUB-2011). We show test samples where our DINO-QPM correctly classifies the image while the dense baseline fails. Columns from left to right: original image~($\boldsymbol{X}$), GradCAM activation map of the dense model, the most similar training sample from the dense-predicted class alongside its cosine similarity score ($\mathrm{sim} = \max_{s \in \mathcal{S}_{c_{\mathrm{pred}}}} \mathrm{CosSim}(\boldsymbol{f}^{\text{froz}}_{\text{CLS}}(\boldsymbol{X}),\, \boldsymbol{f}^{\text{froz}}_{\text{CLS}}(s))$), and the colour-coded local explanation of DINO-QPM for the true class. Row labels indicate the dense prediction (top) and the DINO-QPM prediction (bottom). The dense model consistently confuses Orchard Orioles with visually similar species by attending to non-discriminative regions such as foliage and branches. In contrast, DINO-QPM correctly localises diverse features strictly on the bird's body, enabling accurate classification despite the visual similarity to other species.}
    \label{fig:correct_b}
\end{figure}
\begin{figure}[!ht]
    \centering
    \includegraphics[width=0.75\textwidth]{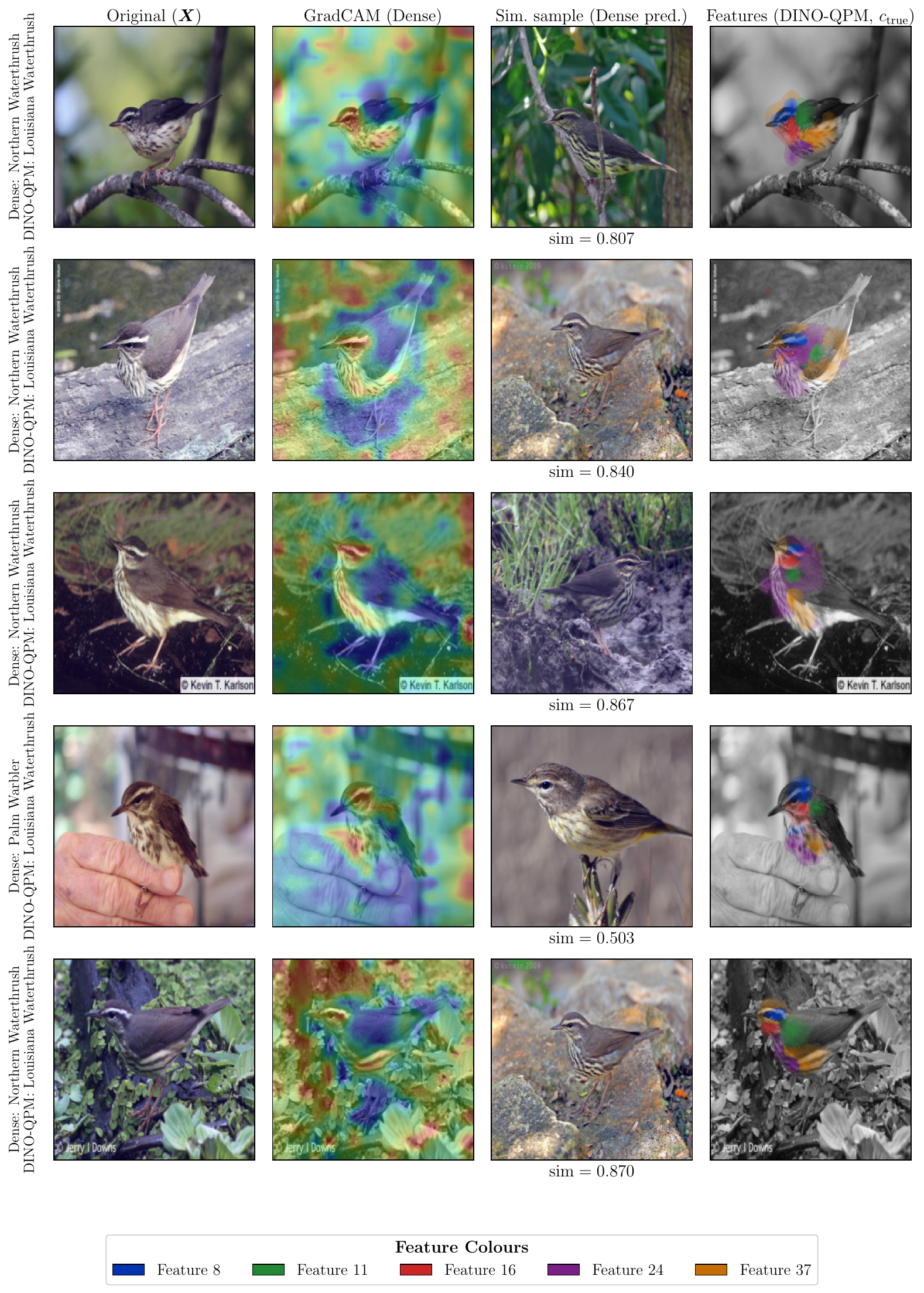}
    \caption{Comparison on the \textit{Louisiana Waterthrush} (CUB-2011). We show test samples where our DINO-QPM correctly classifies the image while the dense baseline fails. Columns from left to right: original image~($\boldsymbol{X}$), GradCAM activation map of the dense model, the most similar training sample from the dense-predicted class alongside its cosine similarity score ($\mathrm{sim} = \max_{s \in \mathcal{S}_{c_{\mathrm{pred}}}} \mathrm{CosSim}(\boldsymbol{f}^{\text{froz}}_{\text{CLS}}(\boldsymbol{X}),\, \boldsymbol{f}^{\text{froz}}_{\text{CLS}}(s))$), and the colour-coded local explanation of DINO-QPM for the true class. Row labels indicate the dense prediction (top) and the DINO-QPM prediction (bottom). The dense model consistently confuses the Louisiana Waterthrush with extremely similar species (e.g., Northern Waterthrush or Palm Warbler), often attending to less discriminative regions. In contrast, DINO-QPM correctly localises diverse features strictly on the bird's body, enabling accurate classification despite the extreme visual overlap between these species.}
    \label{fig:correct_b_2}
\end{figure}

\clearpage
\newpage

\subsection{Dense \texorpdfstring{$\boldsymbol{F}^{\text{froz}}$}{F\textasciicircum{froz}} vs. DINO-QPM Correct Classification}
\begin{figure}[!ht]
    \centering
    \includegraphics[width=\textwidth]{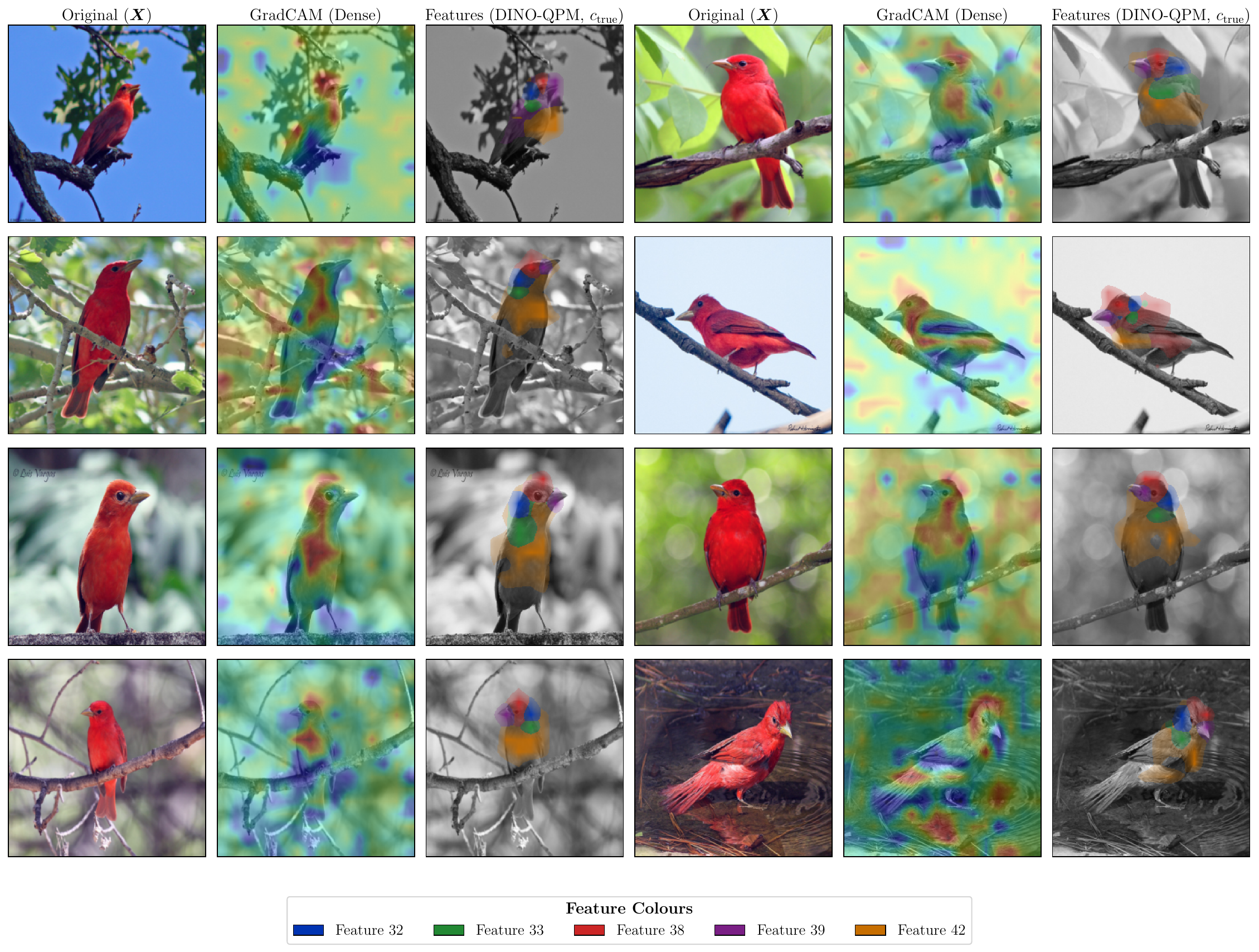}
    \caption{Comparison on the \textit{Summer Tanager} (CUB-2011). We compare the dense baseline and DINO-QPM on eight test images, correctly classified by both models. Each sample is shown as a triplet: the original image~($\boldsymbol{X}$), the GradCAM attribution of the dense model, and the local explanation of DINO-QPM for the true class~$c_{\mathrm{true}}$. The GradCAM attributions of the dense model frequently spread across the background or miss the bird entirely (e.g., samples on the right), demonstrating inconsistent localisation despite correct predictions. In contrast, DINO-QPM's local explanation consistently focuses on the bird, decomposing it into interpretable parts such as the red body plumage (Feature~42), the upper head (Feature~32), and the eye region (Feature~38). This illustrates that DINO-QPM not only localises more reliably but also provides explicit part-level evidence for its decisions, whereas the dense model relies on diffuse, poorly grounded visual cues.}
    \label{fig:correct_both1}
\end{figure}
\begin{figure}[!ht]
    \centering
    \includegraphics[width=\textwidth]{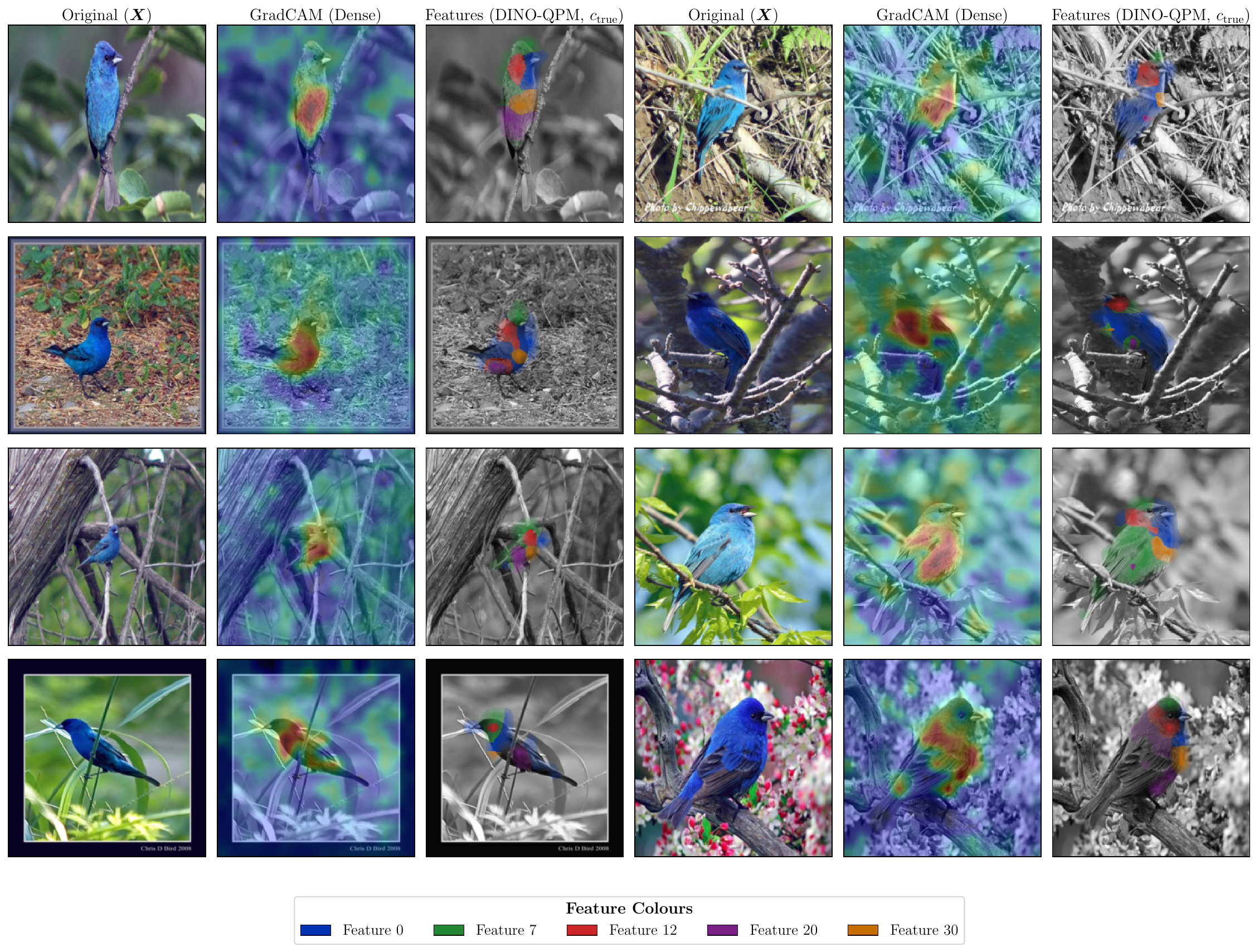}
    \caption{Indigo Bunting samples from the CUB-2011 test set. We compare the dense baseline and DINO-QPM on eight test images, correctly classified by both models. Each sample is shown as a triplet: the original image~($\boldsymbol{X}$), the GradCAM attribution of the dense model, and the local explanation of DINO-QPM for the true class~$c_{\mathrm{true}}$. While both models localise the bird reliably across varying poses and backgrounds, the key difference lies in \emph{what} each model communicates. The dense model focuses on a single discriminative region, resulting in a non-diverse localisation. In contrast, DINO-QPM decomposes its focus into semantically distinct parts---e.g.\ the belly (Feature~20), the mantle and back (Feature~7), and the head region (Feature~12)---offering a richer, more diverse, part-level explanation of \emph{why} the prediction is made.}
    \label{fig:correct_both2}
\end{figure}

\clearpage
\newpage

\subsection{DINO-QPM Failure Analysis}
\begin{figure}[!ht]
    \centering
    \includegraphics[width=0.85\textwidth]{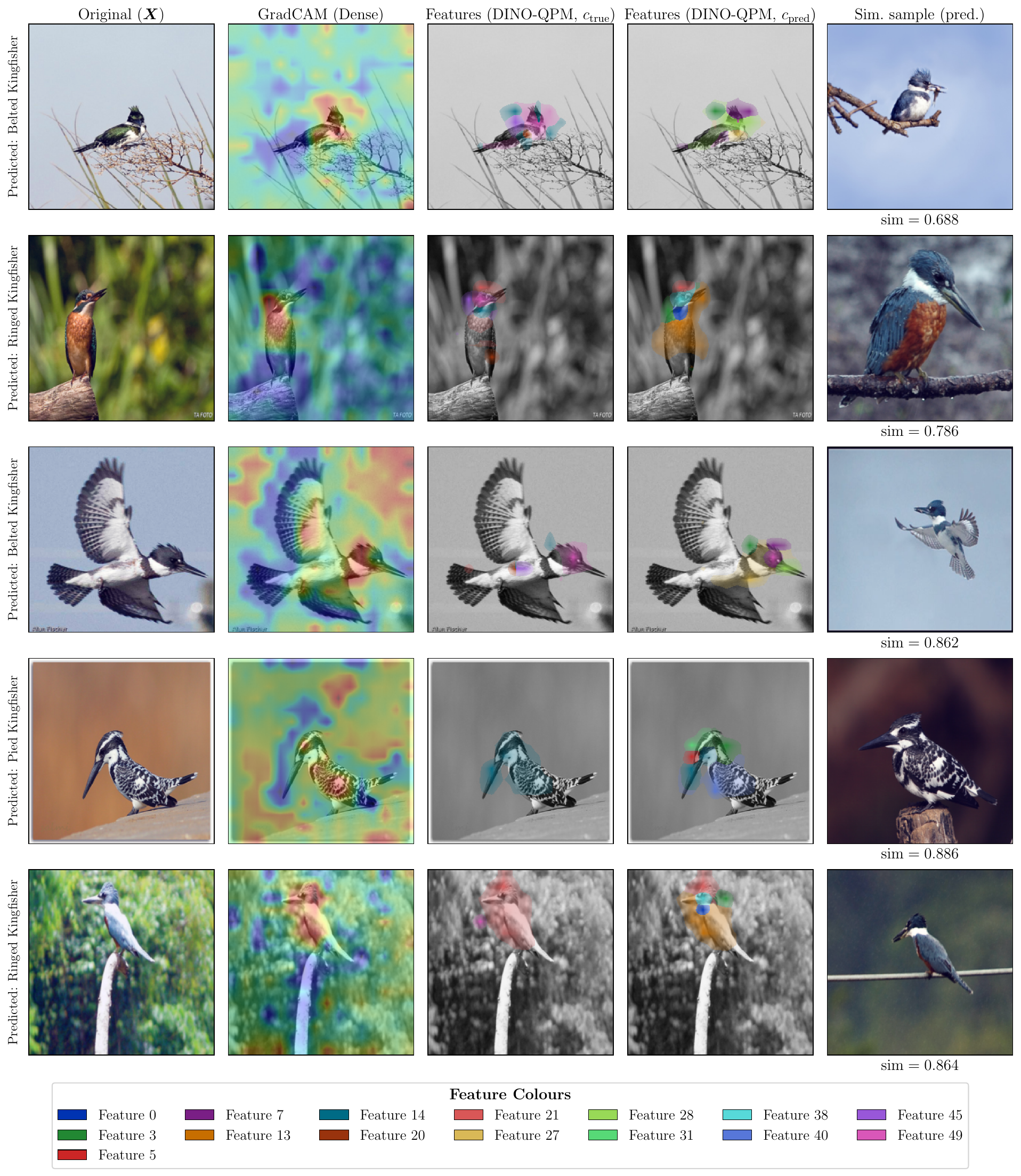}
    \caption{Failure analysis on the \textit{Green Kingfisher} (CUB-2011). Comparison of test samples misclassified by both models. Columns (left to right): original image~($\boldsymbol{X}$); dense model GradCAM; DINO-QPM local explanations for both the true and predicted classes; and the nearest training sample from the predicted class with its cosine similarity ($\mathrm{sim} = \max_{s \in \mathcal{S}_{c_{\mathrm{pred}}}} \mathrm{CosSim}(\boldsymbol{f}^{\text{froz}}_{\text{CLS}}(\boldsymbol{X}),\, \boldsymbol{f}^{\text{froz}}_{\text{CLS}}(s))$). Although both models struggle to distinguish these highly fine-grained kingfisher species, their failure modes differ significantly. The dense model provides no meaningful insight into its errors, whereas DINO-QPM transparently communicates the source of its confusion through faithful, part-level local explanations. Notably, some of these misclassifications might be due to incorrect ground-truth labels \cite{7298658}. For example, the fourth sample appears to be incorrectly annotated, demonstrating how our concept-based explanations can assist in auditing dataset quality.}
    \label{fig:fail_79}
\end{figure}

\begin{figure}[!ht]
    \centering
    \includegraphics[width=0.9\textwidth]{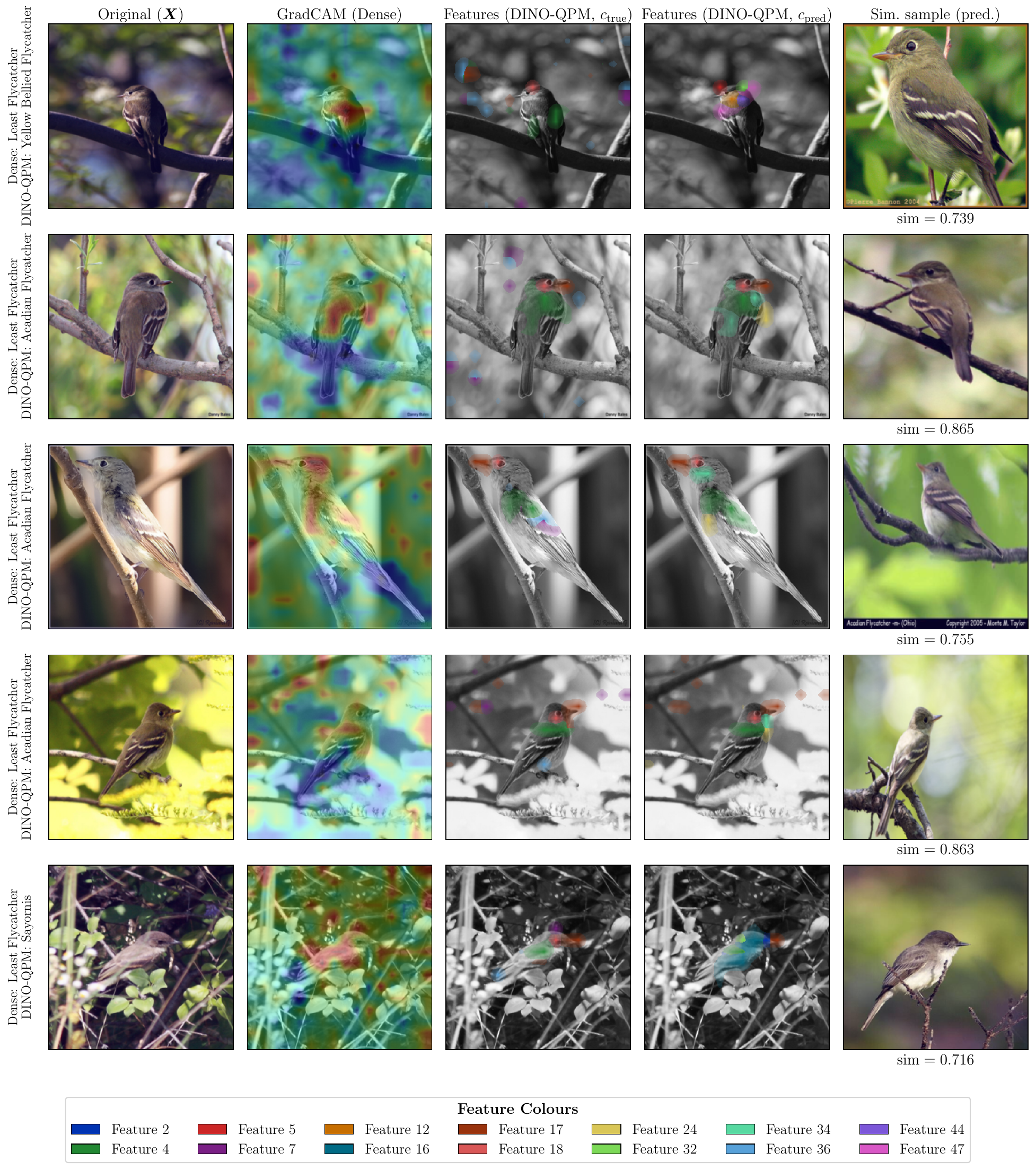}
    \caption{Failure analysis on the \textit{Least Flycatcher} (CUB-2011). We show test samples where the dense baseline classifies correctly but DINO-QPM does not. Columns from left to right: original image~($\boldsymbol{X}$), GradCAM attribution of the dense model, DINO-QPM local explanations for the true and predicted classes, and the most similar training sample from the predicted class with its cosine similarity score ($\mathrm{sim} = \max_{s \in \mathcal{S}_{c_{\mathrm{pred}}}} \mathrm{CosSim}(\boldsymbol{f}^{\text{froz}}_{\text{CLS}}(\boldsymbol{X}),\, \boldsymbol{f}^{\text{froz}}_{\text{CLS}}(s))$). While we previously demonstrated that DINO-QPM successfully overcomes poorly localised dense representations, this strict feature decomposition can occasionally induce errors. In this failure case, although the dense baseline correctly predicts the target class using ungrounded cues, DINO-QPM's refusal to exploit these uninterpretable shortcuts leads to confusion among visually similar flycatcher and \textit{Sayornis} species.}
    \label{fig:fail_flycatcher}
\end{figure}

\clearpage
\newpage

\section{Detailed Results}
\label{app:detailed_results}

\begin{table}[htbp]
    \centering
    \rotatebox{90}{
        \begin{minipage}{0.85\textheight} 
            \centering
            \renewcommand{\arraystretch}{1.1}
            \resizebox{\textwidth}{!}{
                \begin{tabular}{l c cc c cc cc cc}
                    \toprule
                    \multirow{2}{*}{Method} & \multirow{2}{*}{\begin{tabular}{c} Local. \\ Features \end{tabular}} & \multicolumn{2}{c}{Accuracy $\uparrow$} & \multirow{2}{*}{Faithful. $\uparrow$} & \multicolumn{2}{c}{SID@5 $\uparrow$} & \multicolumn{2}{c}{Class-Indep. $\uparrow$} & \multicolumn{2}{c}{Contrast. $\uparrow$} \\
                    \cmidrule(lr){3-4} \cmidrule(lr){6-7} \cmidrule(lr){8-9} \cmidrule(lr){10-11}
                     & & CUB & CARS & & CUB & CARS & CUB & CARS & CUB & CARS \\
                    \midrule
                    DINOv2 $\boldsymbol{f}_{\text{CLS}}^{\text{froz}}$ Linear Probe & \redmark & $\underline{87.9} \pm 0.1$ & $91.7 \pm 0.1$ & $42.6 \pm 0.2$ & $50.9 \pm 0.2$ & $51.5 \pm 0.1$ & $\textbf{99.2} \pm 0.0$ & $\textbf{99.1} \pm 0.0$ & $59.2 \pm 0.0$ & $60.9 \pm 0.0$ \\
                    Dense $\boldsymbol{F}^{\text{froz}}$ & \greenmark & $78.1 \pm 0.3$ & $92.9 \pm 0.1$ & $32.7 \pm 0.2$ & $\textbf{91.8} \pm 0.7$ & $\textbf{93.1} \pm 0.1$ & $\underline{98.8} \pm 0.0$ & $\underline{98.7} \pm 0.0$ & $84.5 \pm 0.3$ & $82.8 \pm 0.1$ \\
                    Resnet50 Baseline \cite{norrenbrockQPMDiscreteOptimization2025} & \greenmark & $83.9 \pm 0.4$ & $92.5 \pm 0.2$ & $60.7 \pm 0.2$ & $57.1 \pm 0.4$ & $51.5 \pm 0.2$ & $98.0 \pm 0.0$ & $97.9 \pm 0.0$ & $74.6 \pm 0.1$ & $75.1 \pm 0.1$ \\
                    \midrule
                    Resnet50 QPM \cite{norrenbrockQPMDiscreteOptimization2025} & \greenmark & $82.9$ & $92.1 \pm 0.2$ & $82.9$ & $89.6$ & $88.2 \pm 0.5$ & $96.8$ & $97.8 \pm 0.0$ & $93.6$ & $\underline{97.1} \pm 0.2$ \\
                    DINO-SLDD & \greenmark & $84.6 \pm 0.4$ & $92.9 \pm 0.1$ & $78.0 \pm 0.9$ & $88.7 \pm 0.3$ & $90.9 \pm 0.8$ & $94.4 \pm 0.1$ & $93.9 \pm 0.2$ & $93.0 \pm 0.3$ & $94.9 \pm 0.5$ \\
                    DINO-QSENN & \greenmark & $85.4 \pm 0.5$ & $\underline{93.3} \pm 0.1$ & $86.0 \pm 0.4$ & $\underline{91.5} \pm 0.5$ & $\underline{92.6} \pm 0.4$ & $93.6 \pm 0.4$ & $94.0 \pm 0.1$ & $\underline{94.4} \pm 0.3$ & $94.9 \pm 0.1$ \\
                    \midrule
                    \midrule
                    DINO-QPM (Ours) & \greenmark & $\textbf{88.3} \pm 0.3$ & $\textbf{94.0} \pm 0.2$ & $\textbf{95.0} \pm 0.6$ & $90.1 \pm 0.0$ & $91.7 \pm 0.2$ & $93.7 \pm 0.1$ & $93.7 \pm 0.1$ & $\textbf{100.0} \pm 0.0$ & $\textbf{100.0} \pm 0.0$ \\
                    DINO-QPM Compact (Ours) & \greenmark & $\textbf{88.3} \pm 0.3$ & $\textbf{94.0} \pm 0.1$ & $\underline{94.4} \pm 0.6$ & $-$ & $-$ & $93.8 \pm 0.1$ & $93.6 \pm 0.1$ & $\textbf{100.0} \pm 0.0$ & $\textbf{100.0} \pm 0.0$ \\
                    \bottomrule
                \end{tabular}
            }
            \vspace{0.5em}
            \caption{Comparison with state-of-the-art interpretable models. We report Accuracy, Faithfulness, SID@5, Class-Independence, and Contrastiveness (all metrics in $\%$). Features of a model are localised if they have a direct connection to the feature vector used for classification. The Faithfulness metric is evaluated only on CUB-2011 due to the availability of segmentation masks. Dense $\boldsymbol{F}^{\text{froz}}$ is the dense model of DINO-QPM and DINOv2 $\boldsymbol{f}_{\text{CLS}}^{\text{froz}}$ Linear Probe is a linear probe \cite{chenSimpleFrameworkContrastive2020} trained on top of the frozen \texttt{CLS} representation. For DINO-SLDD and DINO-QSENN, we employ a pipeline closely resembling the one described in \cref{sec:method}, with the exception of the feature selection mechanisms, which follow \citet{norrenbrockTake5Interpretable2023} and \citet{norrenbrockQSENNQuantizedSelfExplaining2024}, respectively. For Resnet50 QPM \cite{norrenbrockQPMDiscreteOptimization2025} on CUB-2011 we cannot provide standard deviation, as we use the original model provided by authors (\url{https://github.com/ThomasNorr/Qpm}).}
            \label{tab:comparison_sota_landscape} 
        \end{minipage}
    }
\end{table}